\documentclass[preprint,12pt]{elsarticle}

\usepackage[figuresright]{rotating}
\usepackage{amssymb}
\usepackage{amsthm}
\usepackage{multicol}
\usepackage{graphicx}
\usepackage{epstopdf}
\usepackage{fullpage}
\usepackage{latexsym,amsmath}
\usepackage{stmaryrd}
\usepackage{algorithm,algorithmic}
\usepackage{subfigure}
\usepackage{amsfonts}
\usepackage{xcolor,multirow}
\usepackage{appendix}
\usepackage{bm}
\usepackage{url}
\usepackage{diagbox}
\usepackage{array}
\usepackage{booktabs}
\usepackage{tabularx}
\usepackage{footmisc} 
\usepackage{pifont}
\usepackage{tikz}
\usepackage{placeins} 
\usepackage{float}

\usepackage[
justification=justified,
singlelinecheck=true
]{caption}
\newcommand*{\circled}[1]{\lower.7ex\hbox{\tikz\draw (0pt, 0pt)%
		circle (.5em) node {\makebox[1em][c]{\small #1}};}}

\renewcommand{\arraystretch}{1.25}
\renewcommand{\algorithmicrequire}{\textbf{Input:}}
\renewcommand{\algorithmicensure}{\textbf{Output:}}

\newtheorem{theorem}{Theorem}
\newtheorem{definition}{Definition}

\begin{document}
\begin{frontmatter}
		
\title{Quaternion Tensor Modeling for Joint Color--Polarization Demosaicking}

\author[lab1]{Yanqing Song}
\ead{songynu2025@163.com}

\author[lab1]{Jifei Miao\corref{cor1}}
\ead{jifmiao@163.com}

\author[lab1]{Chaoqian Li}
\ead{lichaoqian@ynu.edu.cn}

\author[lab1]{Rui Mei}
\ead{meirui0728@foxmail.com}

\author[lab2]{Kit Ian Kou}
\ead{kikou@umac.mo}

\author[lab3]{Liqiao Yang}
\ead{liqiaoyoung@163.com}

\address[lab1]{School of Mathematics and Statistics, Yunnan University, Kunming, Yunnan, 650091, China}
\address[lab2]{Department of Mathematics, Faculty of
	Science and Technology, University of Macau, Macau 999078, China}
\address[lab3]{School of Computing and Artificial Intelligence
	Southwestern University of Finance and Economics
	Chengdu, Sichuan, China}

\cortext[cor1]{Corresponding author}

\begin{abstract}
	Division-of-focal-plane (DoFP) color polarization cameras enable snapshot acquisition of color polarization mosaic images, but the inherently sparse sampling pattern makes color polarization demosaicking severely ill-posed. Existing methods often fail to jointly exploit the correlations among polarization channels and the physical constraints inherent in polarization imaging, resulting in noticeable demosaicking artifacts. To address this issue, a quaternion-tensor-based color polarization demosaicking (CPDM) method incorporating Stokes-domain total variation (TV) regularization is proposed. Correlation analysis shows that the correlations among polarization channels are stronger than those among color channels. Accordingly, the color polarization images acquired at $0^\circ$, $45^\circ$, $90^\circ$, and $135^\circ$ are encoded into the four components of a third-order quaternion tensor, with the color channels organized along its third mode. A low-rank prior is then imposed on the quaternion tensor to exploit the global structural redundancy in the color polarization data. Moreover, spatial gradients are mapped to the Stokes domain through an orthogonal transformation to separate intensity, polarization and residual variations, with adaptive quaternion weights enabling component-specific regularization and preserving the energy consistency of the reconstructed Stokes vectors. An efficient optimization algorithm is derived for the resulting model. Extensive experiments demonstrate the superior demosaicking performance of the proposed method.
\end{abstract}

\begin{keyword}
	Quaternion tensor\sep color polarization demosaicking\sep polarization-channel correlation\sep low-rank approximation\sep Stokes-domain total variation
\end{keyword}

\end{frontmatter}

\section{Introduction}

Light can be characterized in terms of its intensity, wavelength, polarization, and coherence, whereas conventional color imaging mainly records intensity and spectral information \cite{born2013principles}. Polarization describes the preferred orientations of the electric-field oscillations of light, and polarimetric imaging captures this additional dimension to characterize the polarization state \cite{tyo2006review}. In particular, the polarization information contained in light reflected from the surface of an object provides valuable cues about intrinsic surface properties, including material characteristics and surface geometry, such as surface normals \cite{zhao2024polarization}. However, snapshot division-of-focal-plane (DoFP) polarization cameras typically integrate a micro-polarizer array directly onto the image sensor, such that each pixel records the light intensity transmitted through only one of four linear analyzer orientations (\emph{e.g.}, $0^\circ$, $45^\circ$, $90^\circ$, and $135^\circ$) \cite{xu2022polarization}. To recover complete information, missing angles are estimated via demosaicing methods using spatial neighbors, yielding full resolution images widely utilized in frontier fields, such as underwater image dehazing \cite{wei2021enhancement,wang2024underwater}, road detection \cite{li2021illumination}, biomedical imaging \cite{song2024microstructural}, image fusion \cite{tong2025mspfusion, gao2026dptafusion}, and 3D reconstruction \cite{ngo2022surface}.

\begin{figure}[h]
	\centering
	\includegraphics[width=0.6\linewidth, height=0.35\linewidth]{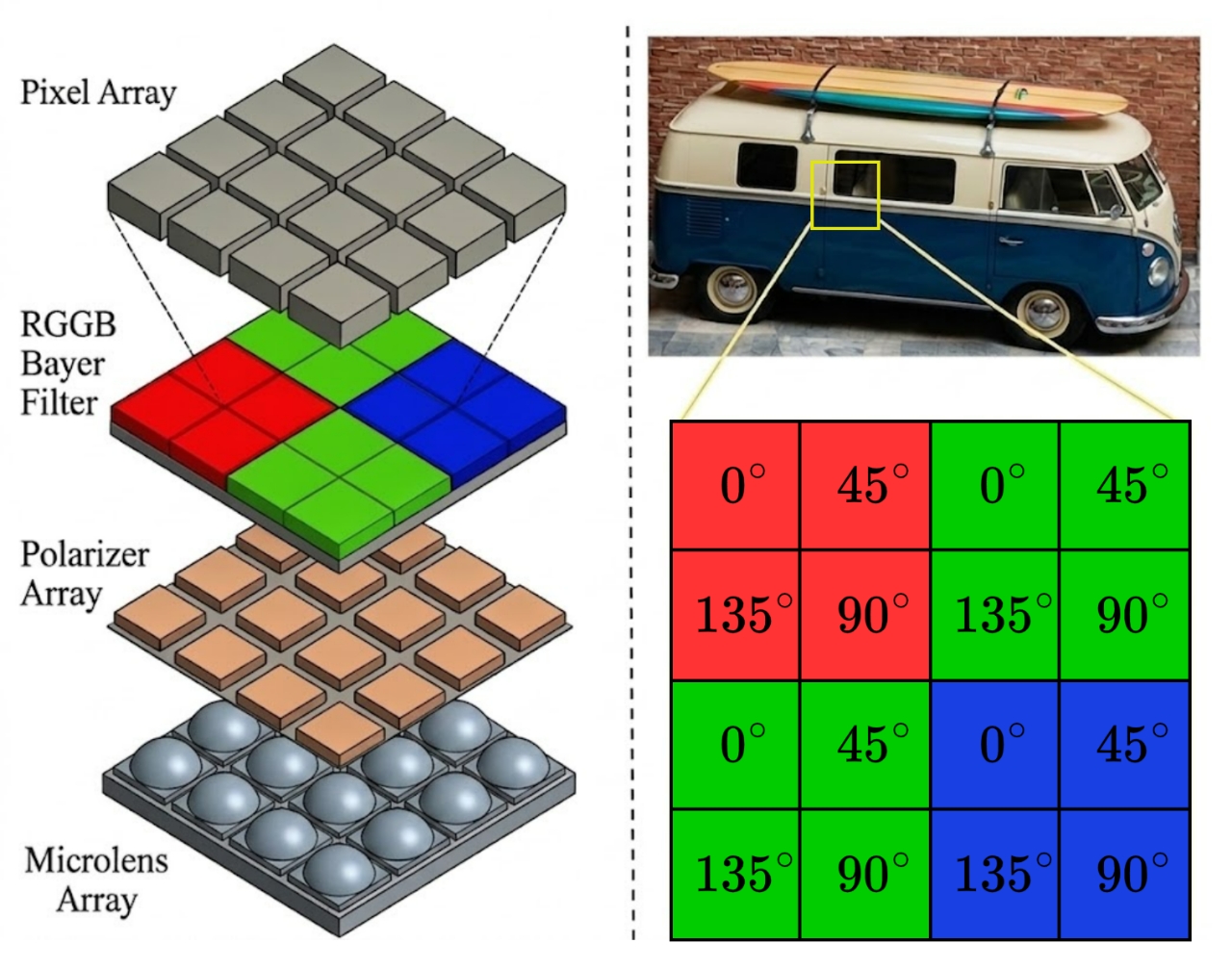} 
	\caption{Schematic of the CPFA architecture and color-polarization mosaic pattern.}
	\label{fig:mosaic}
\end{figure}

Advances in DoFP technology have enabled compact, real-time polarimetric imaging through the integration of a periodic $2\times2$ micro-polarizer array directly onto a monochrome image sensor \cite{xu2022polarization}. Such a design trades spatial resolution for synchronous polarization acquisition in dynamic scenes. To overcome the limitations of monochrome imaging and enable the simultaneous acquisition of color and polarimetric information, the DoFP mechanism has been extended to the more complex color polarization filter array (CPFA). As illustrated in Figure ~\ref{fig:mosaic}, the CPFA jointly captures spatial, spectral, and polarization information by integrating RGB filters with four directional micro-polarizers ($0^\circ$, $45^\circ$, $90^\circ$, and $135^\circ$) within each $4\times4$ super-pixel unit \cite{Bazhyna2006}.
Nevertheless, under the CPFA sampling mechanism, each pixel captures only one of the twelve latent components (three color channels $\times$ four polarization orientations), resulting in extremely sparse observations \cite{luo2023sparse}. Therefore, high-quality color polarization demosaicking (CPDM) methods are essential for reconstructing full-resolution color polarization information and enabling the practical use of CPFA imaging systems.

	\begin{figure}[h]
		\centering
			\begin{minipage}[c]{1\linewidth}
			\centering
			\includegraphics[width=\linewidth]{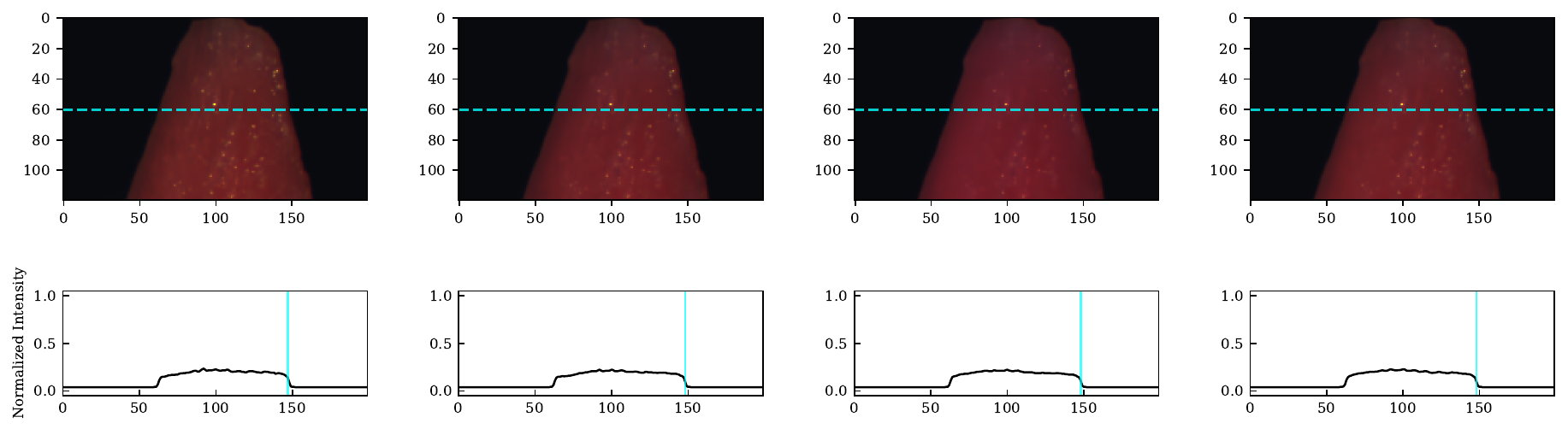}
			\small(a) Profiles of the first sample region.
			\label{fig:region1}   
		\end{minipage}
		\vspace{1em}   
		\begin{minipage}[c]{1\linewidth}
			\centering
			\includegraphics[width=\linewidth]{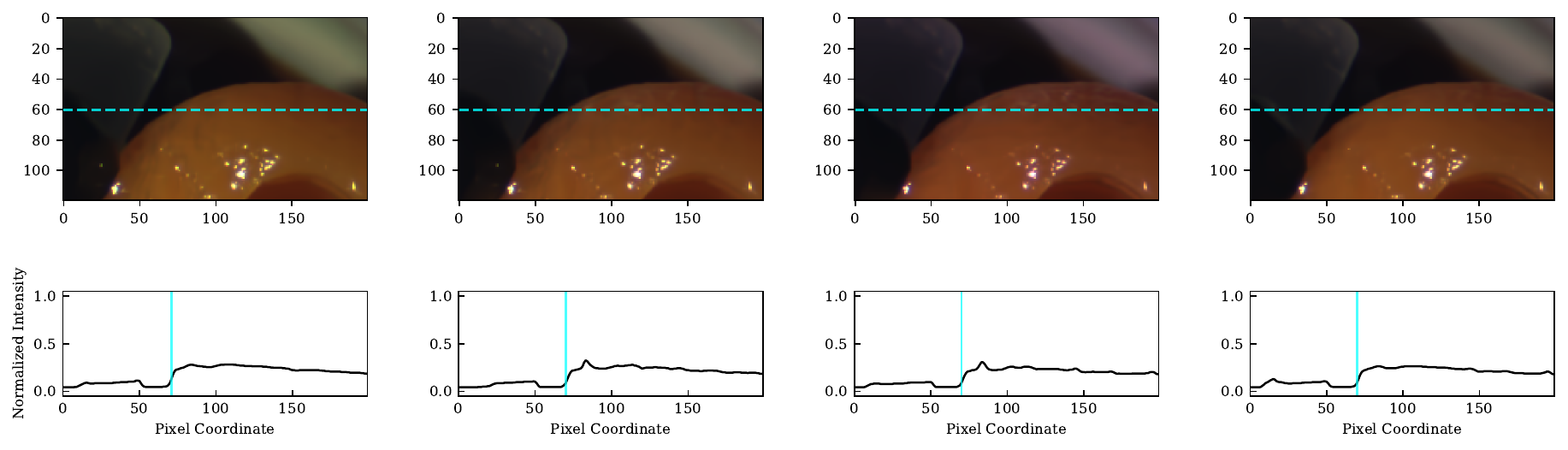}
			\small(b) Profiles of the second sample region.
			\label{fig:region2}   
		\end{minipage}
		\captionsetup{justification=raggedright, singlelinecheck=false}
		\caption{Normalized intensity profiles of four polarization images at
			$0^\circ$, $45^\circ$, $90^\circ$, and $135^\circ$ for two representative
			regions. The cyan dashed lines indicate the horizontal scan lines, and the
			corresponding normalized intensity profiles are shown below. The cyan solid
			lines mark the object--background boundaries, revealing consistent boundary
			locations and similar intensity variations across polarization channels.}
		\label{fig:profiles}
	\end{figure}

Existing CPDM methods can generally be classified into three categories: spatial interpolation methods, optimization-based methods, and deep-learning-based methods. spatial interpolation methods primarily rely on geometric continuity in the spatial domain or channel-wise physical priors. Early studies mainly investigated bilinear, bicubic, and cubic spline interpolation schemes to alleviate instantaneous field of view artifacts in monochrome polarization arrays \cite{Ratliff_LaCasse_Tyo_2009, Gao_Gruev_2011}. However, these linear interpolation operators are prone to severe zipper artifacts and edge degradation. To improve adaptivity, gradient-based interpolation \cite{gao2013gradient}, intensity correlation frameworks \cite{zhang2016image}, and a series of residual interpolation methods  \cite{kim2016four, ahmed2017residual} were subsequently introduced into demosaicking. More recently, methods based on Newton polynomial interpolation with polarization guidance \cite{li2019demosaicking, su2025universal}, as well as approaches incorporating edge aware operators \cite{morimatsu2020monochrome}, polarization difference priors \cite{wu2021polarization}, guided filtering \cite{liu2020new}, hybrid confidence refinement \cite{lu2025hybrid}, and adaptive local gradients \cite{yang2025adaptive}, have been proposed to enhance edge and structural preservation. Nevertheless, these methods still fundamentally rely on local smoothness assumptions, making them prone to zipper artifacts, color distortions, and structural degradation in regions containing complex textures or sharp edges. 

Deep-learning-based methods overcome the performance limitations of traditional interpolation through data-driven modeling capabilities. Since the pioneering polarization convolutional neural network proposed \cite{zhang2018learning}, various network architectures have been successively developed. CPDNet pioneered the end-to-end joint reconstruction framework \cite{Wen_Zheng_Lu_Zhao_2019}. To further suppress interchannel crosstalk, subsequent studies have developed two-stage reconstruction networks \cite{nguyen2022two}, Stokes-guided complementary architectures \cite{zheng2024color}, and active color-perception frameworks \cite{li2025active}. More recently, Li et al.\cite{Li_Luo_Zhang_Yang_2026} incorporate text to image diffusion priors into CPDM task to improve structural fidelity. Despite achieving impressive quantitative performance, these supervised learning methods remain highly dependent on large scale synthetic training datasets. When confronted with unseen real world scenes, their reconstruction robustness deteriorates, frequently resulting in color artifacts and boundary blurring. 

Optimization-based methods  formulate CPDM task as an inverse problem by incorporating spatial priors. Linear minimum mean square error method \cite{dumoulin2022impact} leverages a pre-trained covariance prior to estimate a reconstruction matrix; however, it relies heavily on statistical priors. Mismatched prior statistics will cause unreliable polarization orientation estimation and noticeable boundary distortion. In contrast, Qiu et al.\cite{qiu2021linear} solve the Stokes vector independently within each color channel without modeling the correlation between color information and the corresponding polarization components. Sparse representation-based approaches achieve full  resolution reconstruction via overcomplete dictionary learning \cite{zhang2018sparse}. Wen et al.\cite{wen2021sparse} pre-train distinct overcomplete chromatic and polarimetric dictionaries on a real RGB-polarization dataset; nevertheless, its performance is influenced by the quality of dictionary learning and the distribution training data. Subsequently, Luo et al.\cite{luo2023sparse} further integrate adaptive sub-dictionaries with nonlocal self similarity constraints to enhance structural preservation and denoising capability. More recently, they introduce an online convolutional sparse representation framework for CPDM \cite{luo2024learning}. However, existing optimization-based methods still face several limitations. They often rely on channel-wise independent processing or separately designed priors, leading to relatively high computational costs while failing to fully exploit the latent correlations and physical coupling among the polarization channels. As a result, the complementary information shared across different polarization orientations is not sufficiently utilized, which may cause noticeable demosaicking artifacts and reduced polarization fidelity.
  \begin{figure}[h]
  	\centering
  	\begin{minipage}[c]{0.8\linewidth}   
  		\centering
  		\includegraphics[width=\linewidth]{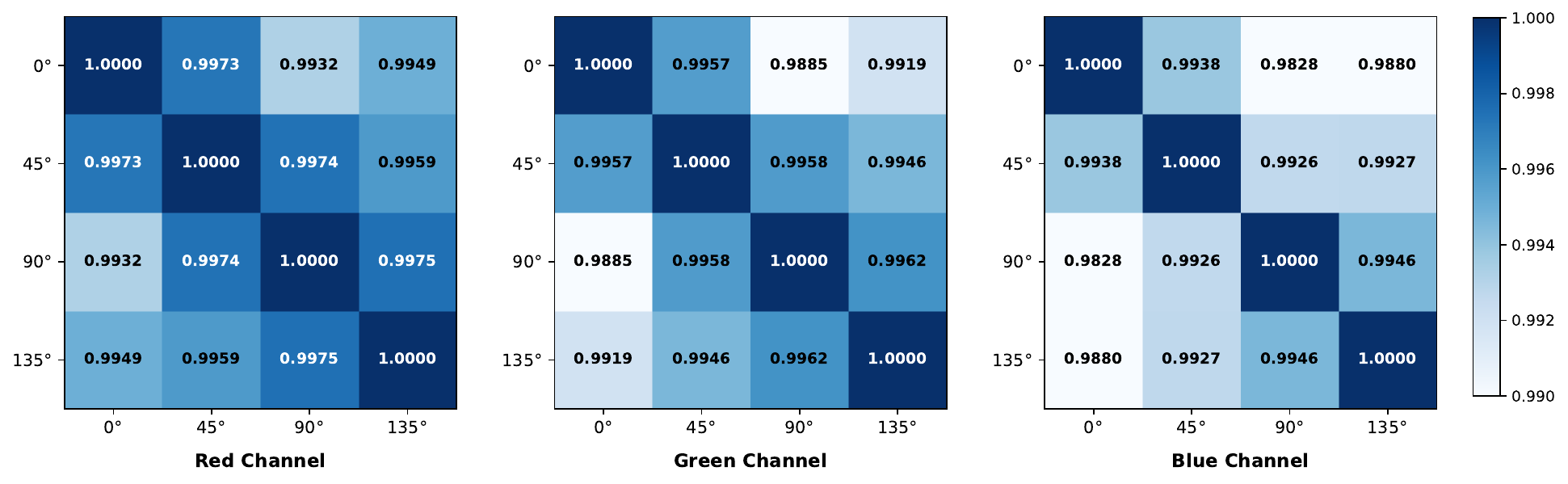}
  		\vspace{2pt} 
  		\begin{minipage}{\linewidth}
  			\small
  			\raggedright
  			(a)  Correlations among polarization channels within each color channel.
  		\end{minipage}
  		\label{fig:polar_heatmap}   
  	\end{minipage}
  	\vspace{1em}   
  	\begin{minipage}[c]{1\linewidth}
  		\centering
  		\includegraphics[width=\linewidth]{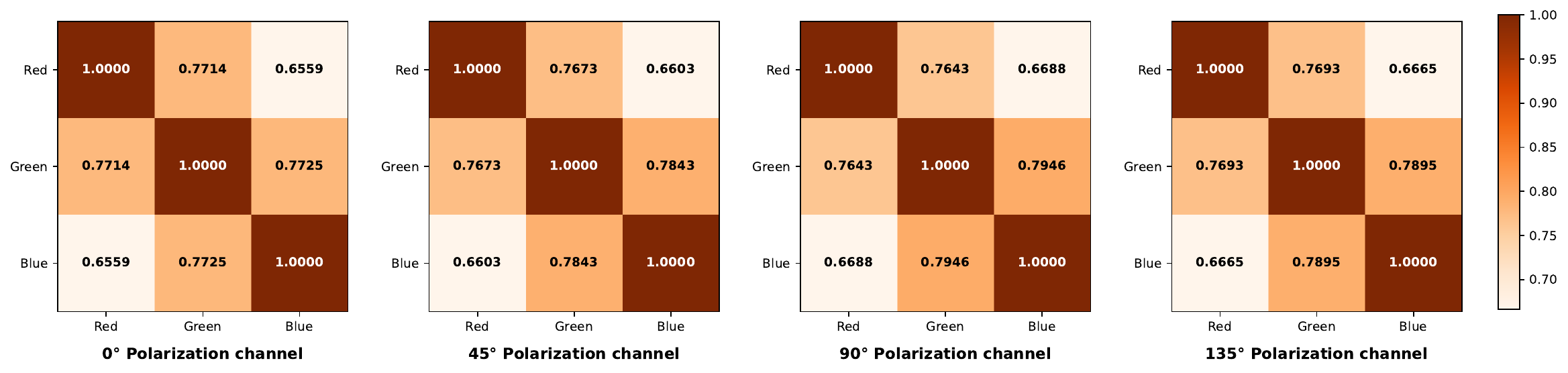}
  		\small (b) Correlations among color channels at each polarization angle.
  		\label{fig:color_heatmap}   
  	\end{minipage}
  	\caption{Pearson correlation analysis of color polarization images. 
  		The polarization channels exhibit substantially stronger correlations than the color channels.}
  	\label{fig:combined}
  \end{figure}
  
To better understand the intrinsic dependencies in color polarization data, we first compare the correlations among polarization channels with those among color channels. As illustrated in Figures \ref{fig:profiles} and \ref{fig:combined}, polarization channels possess far stronger correlations than color channels, with detailed discussions provided in Section \ref{sec:analysis}. To take advantage of this polarization channels correlation, the quaternion tensor \cite{miao2020low} model proposed in this paper embeds $0^\circ$, $45^\circ$, $90^\circ$, and $135^\circ$ color polarization images  into the four components of a quaternion, and treats color as the third dimension of the tensor. This modeling approach maintains a highly coupled structure between color and polarization information. Subsequently, a low-rank regularizer is imposed on the quaternion tensor to characterize its global low-rank structure across the color and polarization dimensions. More importantly, each polarization-channel image jointly encodes total-intensity and angle-dependent polarization information. Direct regularization of the coupled spatial gradients may therefore oversmooth high-frequency details. To address this issue, the gradients are transformed into the Stokes domain, where the decoupled intensity and polarization components are adaptively regularized using component-wise weights encoded in quaternion form, thereby preserving structural details while suppressing artifacts. In addition, for any incident light beam, the sum of light intensities measured by any pair of orthogonal ideal analyzers equals the total irradiance of the beam. Based on this optical property, the proposed transformation can also separate the residual of the total intensity imbalance for subsequent suppression, thereby ensuring energy conservation.

In summary, the main contributions of this work are as follows.
\begin{itemize}
	\item To better exploit the strong correlations among polarization channels, a novel quaternion-tensor-based optimization model is developed. Specifically, the color polarization images acquired at $0^\circ$, $45^\circ$, $90^\circ$, and $135^\circ$ are encoded as the four components of a unified quaternion tensor, with the color channels arranged along the third mode of each component tensor. A low-rank prior is then imposed to characterize the global low-rank structure across the color and polarization dimensions.
	\item To ensure that the aliased intensity and polarization gradient information in every polarimetric image is decoupled into total intensity and polarization difference components, and that the total-intensity consistency constraint is satisfied, we designed an adaptive quaternion penalty weighted regularization term that maps gradients from the spatial domain to the Stokes domain via an orthogonal transformation matrix which preserves the signal energy. 
	\item An efficient ADMM-based algorithm is developed to solve the resulting optimization model. Extensive comparisons with representative interpolation-based, optimization-based, and deep-learning-based methods demonstrate the superior performance of the proposed CPDM method.
\end{itemize}

 The rest of this paper is organized as follows. Section~\ref{sec:2} presents the notation and preliminaries. The motivation and details of the proposed model are elaborated in  Section~\ref{sec:3}. Section~\ref{sec:4} presents the optimization procedure of the model. The experimental results and analysis are demonstrated in Section~\ref{sec:5}, which validates that our method achieves the better performance. Finally, Section \ref{sec:6} concludes this work.

\section{Notation and Preliminaries}\label{sec:2}

This section first introduces the main notation used throughout the paper and then reviews the relevant fundamental concepts.

\subsection{Notation}

\begin{table}[H]
	\centering
	\caption{Summary of the main notation.}
	\label{tab:notation}
	\small
	\renewcommand{\arraystretch}{1.12}
	\setlength{\tabcolsep}{4pt}
	\begin{tabularx}{\linewidth}{
			>{\centering\arraybackslash}p{0.28\linewidth}
			>{\raggedright\arraybackslash}X
		}
		\toprule
		\textbf{Symbol} & \textbf{Description} \\
		\midrule
		
		$\mathbb{R}$, $\mathbb{C}$, $\mathbb{H}$
		& Sets of real numbers, complex numbers, and quaternions \\
			
		$\dot{a}$, $\dot{\mathbf{a}}$,
		$\dot{\mathbf{A}}$, $\dot{\mathcal{A}}$
		& Quaternion scalar, vector, matrix, and tensor, respectively \\
		
		$\dot{\mathcal{A}}(:,:,i_{3},\ldots,i_{N})$
		& Matrix slice obtained by fixing the indices of modes $3,\ldots,N$ \\
		
		$\dot{\mathcal{A}}^{(i)}$
		& Compact notation for the corresponding matrix slice \\
		
		$(\cdot)^{T}$, $(\cdot)^{H}$
		& Transpose and conjugate transpose \\
		
		$(\cdot)^{*}$, $(\cdot)^{-1}$
		& Quaternion conjugate and inverse \\
		
		$|\cdot|$,
		$\|\cdot\|_{F}$
		& Modulus and frobenius norm\\
		
		 $\|\cdot\|_{w,*}$,
		 $\|\cdot\|_{1}$
		& Weighted nuclear norm and $\ell_1$-norm  \\
		
		$\Vert{} \cdot \Vert{}_{1, \mathrm{comp}}$ 
		& Component-wise $\ell_1$-norm of a quaternion tensor\\
		
		$\nabla_{d}$
		& Gradient operator along the $d$th spatial direction \\
		
		$\operatorname{stack}(\cdot)$& Vertical concatenation of the arguments\\
		
		$(\cdot)_{+}$& Positive-part operator\\
		
		$\odot$, $\oslash$
		& Element-wise product and element-wise division\\
		
		$\times_{k}$
		& Mode-$k$ product \\
		
		$\times_{\mathrm{comp}}$
		& Transformation along the quaternion-component mode \\
		
		$\odot_{\mathrm{comp}}$ 
		& Component-wise multiplication between the quaternion components
		\\
		$\ast$ 
		& Spatial convolution\\
		\bottomrule
	\end{tabularx}
\end{table}

\subsection{Preliminaries}
The preliminaries underlying the proposed method are presented in two parts: quaternion fundamentals and Stokes parameter fundamentals.

\subsubsection{Quaternion Fundamentals}
As a high dimensional extension of complex numbers, quaternions were first proposed by Hamilton in 1843 \cite{hamilton1844quaternions}. In the quaternion domain $\mathbb{H}$, an arbitrary $N$th-order quaternion tensor $\dot{\mathcal{T}} \in \mathbb{H}^{I_1 \times I_2 \times \dots \times I_N}$ consists of one real part and three imaginary parts \cite{miao2020low}, formulated as
\begin{equation}\dot{\mathcal{T}} = \mathcal{T}_0 + \mathcal{T}_1 i + \mathcal{T}_2 j + \mathcal{T}_3 k,\end{equation}
where $\mathcal{T}_n \in \mathbb{R}^{I_1 \times I_2 \times \dots \times I_N}$ ($n\in\{0,1,2,3\}$) are real-valued tensors of the same size. Specifically, $\dot{\mathcal{T}}$ degenerates to a quaternion scalar $\dot{q}$ when $I_1=\dots=I_N=1$, and to a quaternion matrix $\dot{\mathbf{Q}}$ when $N=2$. If the real component $\mathcal{T}_0$ is a zero tensor, $\dot{\mathcal{T}}$ is termed a pure quaternion tensor. The imaginary units $i, j, k$ obey the fundamental algebraic rules:
\[
\left\{
\begin{aligned}
	&i^{2}=j^{2}=k^{2}=ijk=-1,\\
	&ij=-ji=k,
	jk=-kj=i,
	ki=-ik=j.
\end{aligned}
\right.
\]
Notably, due to the non-commutative nature of quaternion multiplication, $\dot{p}\dot{q} \neq \dot{q}\dot{p}$ generally holds for arbitrary $\dot{p},\dot{q} \in \mathbb{H}$. The conjugate of a quaternion $\dot{q} = q_0 + q_1 i + q_2 j + q_3 k$ is defined as $\dot{q}^* = q_0 - q_1 i - q_2 j - q_3 k$, which satisfies $(\dot{p}+\dot{q})^* = \dot{p}^*+\dot{q}^*$ and $(\dot{p}\dot{q})^* = \dot{q}^*\dot{p}^*$. The norm (modulus) of a quaternion is defined as $|\dot{q}| = \sqrt{\dot{q}\dot{q}^*} = \sqrt{q_0^2 + q_1^2 + q_2^2 + q_3^2}$.
\begin{definition}[$\star_{QT}$-Product \cite{miao2023quaternion}]
	Given two quaternion tensors $\dot{\mathcal{A}} \in \mathbb{H}^{I_1 \times l \times I_3 \times \dots \times I_N}$ and $\dot{\mathcal{B}} \in \mathbb{H}^{l \times I_2 \times I_3 \times \dots \times I_N}$, together with a set of invertible quaternion matrices $\{\dot{\mathbf{Q}}_n\in \mathbb{H}^{I_n \times I_n}\}_{n=3}^N$, the $\star_{QT}$-product is defined as
	\begin{equation}
		\dot{\mathcal{T}} = \dot{\mathcal{A}} \star_{QT} \dot{\mathcal{B}} = ( \hat{\dot{\mathcal{A}}} \star_{QF} \hat{\dot{\mathcal{B}}} ) \times_3 \dot{\mathbf{Q}}_3^{-1} \times_4 \dots \times_N \dot{\mathbf{Q}}_N^{-1},
	\end{equation}
	where $\hat{\dot{\mathcal{A}}} = \dot{\mathcal{A}} \times_3 \dot{\mathbf{Q}}_3 \times_4 \cdots \times_N \dot{\mathbf{Q}}_N, \hat{\dot{\mathcal{B}}} = \dot{\mathcal{B}} \times_3 \dot{\mathbf{Q}}_3 \times_4 \cdots \times_N \dot{\mathbf{Q}}_N$, $\times_k$ denotes the mode-$k$ product of tensors, and $\star_{QF}$ is the quaternion facewise product, i.e., $\dot{\mathcal{F}} = \hat{\dot{\mathcal{A}}} \star_{QF} \hat{\dot{\mathcal{B}}}$ such that the frontal slice of $\dot{\mathcal{F}}$ satisfies $\dot{\mathcal{F}}(:,:,i_3,\dots,i_N) = \hat{\dot{\mathcal{A}}}(:,:,i_3,\dots,i_N)\hat{\dot{\mathcal{B}}}(:,:,i_3,\dots,i_N)$, $i_k = 1,\dots,I_k$ for $k=3,\dots,N$.
\end{definition}

\begin{definition}[Conjugate transpose \cite{miao2023quaternion}]

	Given an $N$th-order ($N\ge 3$) quaternion tensor $\dot{\mathcal{T}} \in \mathbb{H}^{I_1\times I_2\times \dots\times I_N}$, its conjugate transpose is denoted as $\dot{\mathcal{T}}^H \in \mathbb{H}^{I_2\times I_1\times \dots\times I_N}$, satisfying
	\[
	\hat{\dot{\mathcal{T}}}^H(:,:,i_3,\dots,i_N) = \big(\hat{\dot{\mathcal{T}}}(:,:,i_3,\dots,i_N)\big)^H,
	\]
	where $i_k = 1,\dots,I_k$ for $k=3,\dots,N$.

\end{definition}

\begin{definition}[Identity quaternion tensor \cite{miao2023quaternion}]
	The identity $N$th-order ($N\ge 3$) quaternion tensor $\dot{\mathcal{I}} \in \mathbb{H}^{P\times P\times I_3\times \dots\times I_N}$ is the quaternion tensor satisfying that each frontal slice of $\hat{\dot{\mathcal{I}}}$ is the identity quaternion matrix, i.e.,
	\begin{equation*}
		\hat{\dot{\mathcal{I}}}(:,:,i_3,\dots,i_N) = \dot{\mathbf{I} }\in \mathbb{H}^{P\times P}, \label{eq:identity_qtensor}
	\end{equation*}
	where $\dot{\mathbf{I}}$ represents the identity quaternion matrix, which is the same as the traditional real-valued identity matrix, $i_k = 1,\dots,I_k$ for $k=3,\dots,N$.
\end{definition}
\begin{definition}[Unitary quaternion tensor \cite{miao2023quaternion}]
	An $N$th-order ($N\ge 3$) quaternion tensor $\dot{\mathcal{T}} \in \mathbb{H}^{P\times P\times I_3\times \dots\times I_N}$ is unitary if
	\[
	\dot{\mathcal{T}}^H \star_{QT} \dot{\mathcal{T}} = \dot{\mathcal{I}} = \dot{\mathcal{T}} \star_{QT} \dot{\mathcal{T}}^H.
	\]
\end{definition}

\begin{theorem}[TQt-SVD \cite{miao2023quaternion}]
	Given a quaternion tensor $\dot{\mathcal{T}} \in \mathbb{H}^{I_1 \times I_2 \times \dots \times I_N}$. There exist two unitary quaternion tensors $\dot{\mathcal{U}} \in \mathbb{H}^{I_1 \times I_1 \times \dots \times I_N}$ and $\dot{\mathcal{V}} \in \mathbb{H}^{I_2 \times I_2 \times \dots \times I_N}$ such that
	
	\begin{equation}
		\dot{\mathcal{T}} = \dot{\mathcal{U}} \star_{QT} \dot{\mathcal{D}} \star_{QT} \dot{\mathcal{V}}^H,
	\end{equation}
where $\dot{\mathcal{D}} \in
\mathbb{H}^{I_1\times I_2\times\cdots\times I_N}$
is an f-diagonal quaternion tensor, i.e., each of its frontal
slices is a diagonal quaternion matrix.
\end{theorem}

\begin{definition}[TQt-rank \cite{miao2023quaternion}]\label{def:5}
	Let $\dot{\mathcal{T}} \in \mathbb{H}^{I_1 \times I_2 \times \dots \times I_N}$ ($N \ge 3$), and its corresponding TQt-SVD be $\dot{\mathcal{T}} = \dot{\mathcal{U}} \star_{QT} \dot{\mathcal{D}} \star_{QT} \dot{\mathcal{V}}^H$. The TQt-rank of $\dot{\mathcal{T}}$ is defined as the number of non-zero tubes in $\dot{\mathcal{D}}$, i.e.,
	
	\begin{equation}
		\mathrm{rank}_{TQt}(\dot{\mathcal{T}}) = \#\{ k \mid \| \dot{\mathcal{D}}(k,k,:,\dots,:) \|_F > 0 \},
	\end{equation}
	where $k=1,\ldots,K$ and $K=\min\{I_1,I_2\}$. Moreover, the $k$th singular value of $\dot{\mathcal{T}}$ is defined as
	\begin{equation}
    \sigma_k(\dot{\mathcal{T}}) = \| \dot{\mathcal{D}}(k,k,:,\dots,:) \|_F.
	\end{equation}
\end{definition}

\begin{definition}[Quaternion tensor weighted nuclear norm (QT-WNN) \cite{yu2019quaternion}]
	For a third-order quaternion tensor $\dot{\mathcal{T}} \in \mathbb{H}^{I_1 \times I_2 \times I_3}$,
	the weighted quaternion tensor nuclear norm  is defined as
	\begin{equation}
		\|\dot{\mathcal{T}}\|_{w,*}
		= \sum_{i=1}^{I_3} \sum_{j=1}^{\min\{I_1,I_2\}} w_{i,j} \sigma_j(\dot{\mathcal{T}}^{(i)}),
	\end{equation}
	where $w_{i,j} \ge 0$ is a non-negative weight assigned to $\sigma_j(\dot{\mathcal{T}}^{(i)})$. 
	
\end{definition}

\begin{definition}[Discrete gradient of quaternion tensor \cite{yang2026nonlinear}]
	Let $\dot{\mathcal{T}} \in \mathbb{H}^{n_1 \times n_2 \times n_3}$ be a third-order quaternion tensor. For each mode $d \in \{1, 2, 3\}$, the discrete gradient operator is defined as
	\begin{equation}
		\nabla_d\dot{\mathcal{T}} = \dot{\mathcal{T}} \times_d \mathbf{D}_{n_d},
	\end{equation}
	where $\mathbf{D}_{n_d} \in \mathbb{R}^{n_d \times n_d}$ is a row circulant matrix of $(-w_{n_d}, w_{n_d}, 0, \dots, 0)$, and $w_{n_d}>0$. The full spatial gradient of $\dot{\mathcal{T}}$ is given by the concatenation of gradients along each mode,
	\begin{equation}
		\nabla \dot{\mathcal{T}} = \big(\nabla_1 \dot{\mathcal{T}},\ \nabla_2 \dot{\mathcal{T}},\ \nabla_3 \dot{\mathcal{T}}\big).
	\end{equation}
\end{definition}

\subsubsection{Stokes Parameter Fundamentals}
Polarization imaging captures physical information associated with object surfaces beyond that provided by conventional intensity imaging by measuring the polarization state of light \cite{tyo2006review}. This polarization state is fully characterized by the Stokes vector \cite{azzam2016stokes}
\begin{equation*}
	\mathbf{S}=[S_0,S_1,S_2,S_3]^T,
\end{equation*}
where $S_0$ denotes the total light intensity; $S_1$ and $S_2$ quantify the intensity differences between the horizontal and vertical linear polarization components and between the $45^\circ$ and $135^\circ$ linear polarization components, respectively; and $S_3$ characterizes the circular polarization component \cite{born1999principles,hu2026stokes}. For color polarization sensors that measure only linear polarization, the first three Stokes parameters can be recovered from the intensity measurements acquired at four polarization orientations, namely, $0^\circ$, $45^\circ$, $90^\circ$, and $135^\circ$. Denoting the intensity measured at orientation $\theta$ by $I_{\theta}$, the Stokes parameters are computed as
\begin{equation}\label{Stokesp}
	\begin{aligned}
		S_0 &= I_{0^\circ}+I_{90^\circ}
		= I_{45^\circ}+I_{135^\circ},\\
		S_1 &= I_{0^\circ}-I_{90^\circ},\\
		S_2 &= I_{45^\circ}-I_{135^\circ}.
	\end{aligned}
\end{equation}
Based on the Stokes parameters, the degree of linear polarization (DoLP) is defined as
\begin{equation}
	\mathrm{DoLP}
	=
	\frac{\sqrt{S_1^2+S_2^2}}{S_0},
\end{equation}
which quantifies the proportion of linearly polarized light in the total intensity and provides useful cues about surface reflection and scattering properties. The angle of linear polarization (AoLP) characterizes the orientation of linear polarization relative to a reference axis and serves as an important descriptor of scene geometry and surface properties \cite{zhu2019camera}. It is defined as
\begin{equation}
	\mathrm{AoLP}
	=
	\frac{1}{2}\operatorname{atan2}(S_2,S_1).
\end{equation}

In practical CPDM tasks, the primary focus is on linearly polarized light. Owing to manufacturing constraints, mainstream commercial polarization image sensors, such as the Sony IMX series, typically employ micro-polarizer arrays containing linear polarization filters with transmission axes oriented at $0^\circ$, $45^\circ$, $90^\circ$, and $135^\circ$. Acquiring the circular polarization component $S_3$ requires additional optical elements, such as quarter-wave plates, which substantially increase system complexity, manufacturing difficulty, and spatial overhead \cite{Ratliff_LaCasse_Tyo_2009}. Moreover, light reflected from and scattered by common surfaces generally exhibits extremely weak circular polarization \cite{tyo2006review}. Consequently, neglecting the circular polarization component has become common practice in both academic research and industrial applications.

\section{Motivation and Proposed Model}
\label{sec:3}

This section first introduces the motivation behind the proposed model and then presents its mathematical formulation.

\subsection{Motivation}\label{sec:analysis}

\textbf{Polarization-Channel Correlation}: Polarimetric images exhibit strong correlations among the measurements acquired at $0^\circ$, $45^\circ$, $90^\circ$, and $135^\circ$, motivating the proposed quaternion tensor representation. Pearson correlation analysis conducted by Mihoubi et al. shows that correlations among polarization channels are generally stronger than those among spectral channels \cite{Mihoubi2018}. Similar observations have also been reported in subsequent studies \cite{liu2020new,xin2023demosaicking,yi2024demosaicking,liu2025efficient}, indicating that different polarization channels share highly consistent geometric structures, edges, and textures.

To further illustrate this property, Figure \ref{fig:profiles} presents the normalized intensity profiles of different channels. Their closely aligned variations indicate strong consistency in edge and texture locations. The correlation matrices in Figure \ref{fig:combined} provide additional quantitative evidence. Specifically, Figure \ref{fig:combined}(a) shows strong correlations among the four polarization channels within each RGB channel, whereas Figure \ref{fig:combined}(b) reveals comparatively weaker correlations among the RGB channels under each polarization orientation. These results confirm that the correlations among polarization channels are stronger than those among color channels.

\begin{figure*}[!htbp]
	\centering
	\includegraphics[width=0.8\textwidth]{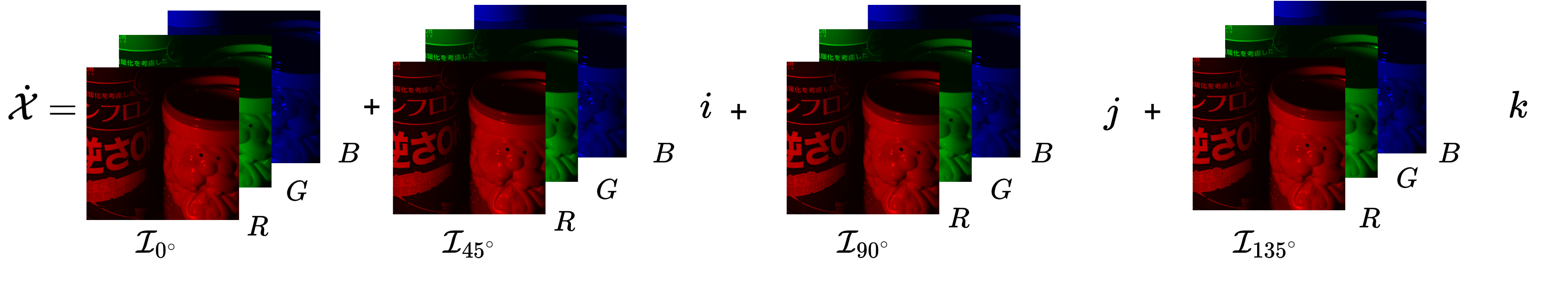}
	\vspace{-0.2cm}
	\caption{Quaternion tensor representation of four color polarization images, with polarization angles encoded as quaternion components and RGB channels arranged along the third mode.}
	\label{fig:quaternion_tensor}
\end{figure*}

Despite  strong correlations exist among polarization channels, many CPDM methods reconstruct them independently, thereby underutilizing their shared structural information. To exploit these correlations, the color polarization images acquired at $0^\circ$, $45^\circ$, $90^\circ$, and $135^\circ$ are assigned to the four components of a third-order quaternion tensor, as illustrated in Figure \ref{fig:quaternion_tensor}. The RGB channels are arranged along the third mode, yielding
\begin{equation}\label{eq:qp}
	\dot{\mathcal{X}}
	=
	\mathcal{I}_{0^\circ}
	+
	\mathcal{I}_{45^\circ} i
	+
	\mathcal{I}_{90^\circ} j
	+
	\mathcal{I}_{135^\circ} k.
\end{equation}

The construction (\ref{eq:qp}) provides a unified and structured representation of the four polarization channels. Rather than treating them as independent real-valued tensors, the quaternion formulation couples the polarization measurements through its four components, allowing their shared spatial structures and cross-channel correlations to be jointly exploited in subsequent tensor operations. Meanwhile, organizing the RGB channels along the third mode preserves the intrinsic spectral structure of the data. Consequently, the proposed representation avoids channel-wise reconstruction and provides a natural foundation for imposing a low-rank quaternion tensor prior.

\textbf{Low-Rankness in the TQt-Rank Sense}: Color polarization images exhibit substantial spatial and cross-channel correlations. Away from object boundaries and material discontinuities, surface normals and material properties generally vary smoothly, leading to strong correlations among neighboring pixels. Moreover, the color and polarization channels describe the same scene and therefore tend to share consistent edges and spatial structures \cite{hu2025}. These characteristics suggest that color polarization images possess an inherent low-rank structure.

\begin{figure*}[h]
	\centering
	\includegraphics[width=0.8\textwidth]{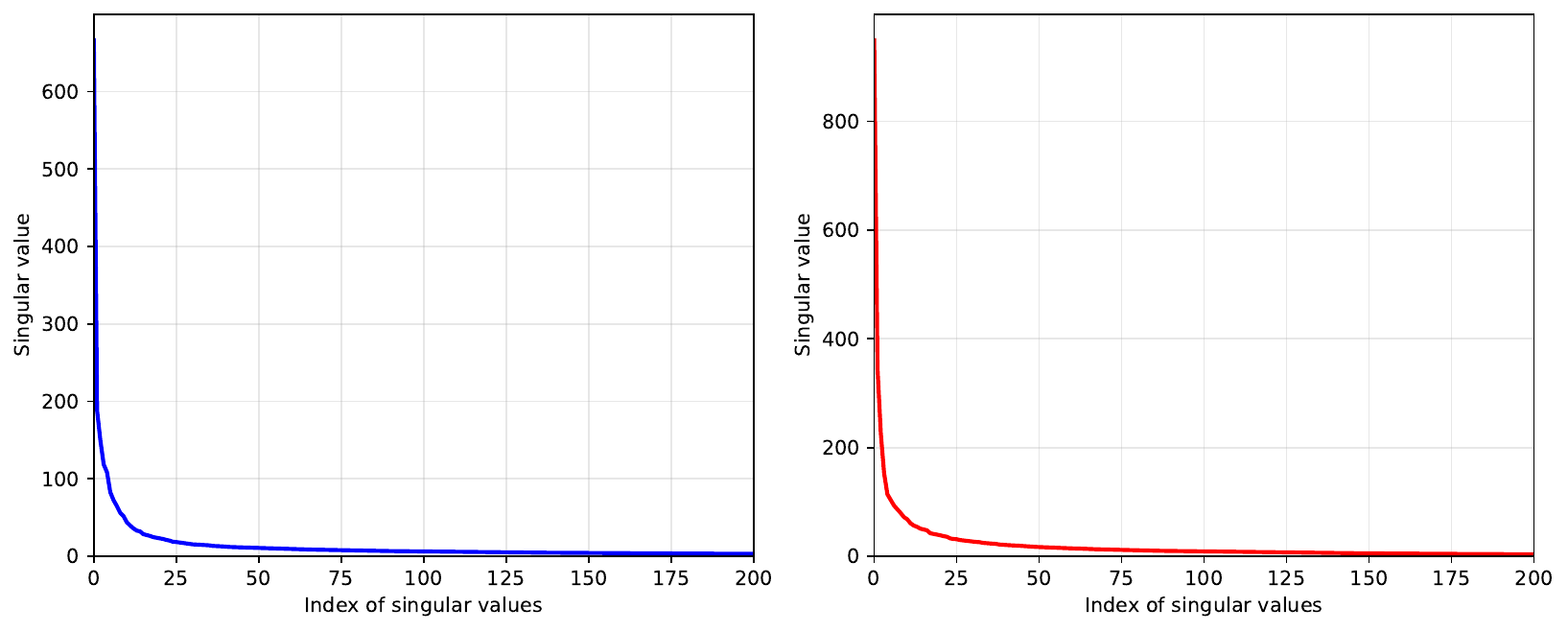}
	\vspace{-0.2cm}
	\caption{Tensor singular value distributions of two representative color polarization images, demonstrating their low-rankness in the TQt-rank sense.}
	\label{fig:singular_values}
\end{figure*}

To verify this property under the adopted quaternion tensor framework, their low-rankness is examined in the TQt-rank sense. Figure~\ref{fig:singular_values} presents the distributions of the tensor singular values defined in Definition~\ref{def:5} for two representative scenes. Most singular values decay rapidly toward zero and are substantially smaller than the first few dominant ones. This observation indicates that color polarization images can be well approximated by low-rank quaternion tensors, thereby motivating the incorporation of a low-rank quaternion tensor prior for recovering the missing structural information.

\textbf{Stokes-Domain Gradient Decoupling and Adaptive Regularization}: Conventional total variation (TV) regularization is typically imposed directly on the four polarization intensity images
$\mathcal{I}_{0^\circ}$, $\mathcal{I}_{45^\circ}$,
$\mathcal{I}_{90^\circ}$, and $\mathcal{I}_{135^\circ}$.
However, in each angular intensity measurement, the total-intensity component is coupled with one of the two linear-polarization components. Specifically, according to (\ref{Stokesp}), the four polarization intensity images can be expressed in a unified form as
\begin{equation}
	\mathcal{I}_{\theta}
	=
	\frac{1}{2}
	\left[
	\mathcal{S}_{0}
	+
	\cos(2\theta)\mathcal{S}_{1}
	+
	\sin(2\theta)\mathcal{S}_{2}
	\right],
	\qquad
	\theta\in
	\left\{
	0^\circ,45^\circ,90^\circ,135^\circ
	\right\}.
	\label{eq:intensity_stokes}
\end{equation}
Since $\theta$ is fixed for each polarization channel, $\cos(2\theta)$ and $\sin(2\theta)$ are constants. Applying the spatial gradient operator to both sides of \eqref{eq:intensity_stokes} and using its linearity yields
\begin{equation}
	\nabla \mathcal{I}_{\theta}
	=
	\frac{1}{2}\nabla \mathcal{S}_{0}
	+
	\frac{1}{2}\cos(2\theta)\nabla \mathcal{S}_{1}
	+
	\frac{1}{2}\sin(2\theta)\nabla \mathcal{S}_{2}.
	\label{eq:gradient_mixture}
\end{equation}
Equation~\eqref{eq:gradient_mixture} shows that the spatial gradient of each polarization intensity image is a linear combination of the total-intensity gradient $\nabla\mathcal{S}_{0}$ and the polarization-difference gradients $\nabla\mathcal{S}_{1}$ and $\nabla\mathcal{S}_{2}$. Therefore, directly applying TV regularization to the angular intensity images cannot distinguish intensity-related structures from polarization-related variations and may oversmooth physically meaningful polarization details. Moreover, when the polarimetric extinction ratio and transmittance of the analyzer are assumed to be ideal, the residual of total-intensity inconsistency between the two orthogonal polarization pairs \((0^\circ,90^\circ)\) and \((45^\circ,135^\circ)\) vanishes at all spatial and color coordinates under such idealized conditions. In realistic imaging scenarios, however, the analyzer is non-ideal; furthermore, sensor noise, lens aberration, polarizer transmittance error and illumination fluctuation together produce non-negligible residual across the whole field of view \cite{yan2022pirc}.

To separate these physical gradient components and suppress the residual while preserving the four-component quaternion representation, an orthogonal Stokes-consistency transformation is introduced. For each gradient direction $d$, define
 $\mathbf{g}_d = \operatorname{stack}(\nabla_d\mathcal{I}_{0^\circ},\allowbreak\nabla_d\mathcal{I}_{45^\circ},\allowbreak\nabla_d\mathcal{I}_{90^\circ},\allowbreak\nabla_d\mathcal{I}_{135^\circ})$,
where $\operatorname{stack}(\cdot)$ denotes the vertical concatenation of its arguments. The proposed transformation matrix is given by\footnote{The choice of $\mathbf{W}$ is not unique. Any orthogonal transformation that achieves the desired separation of the inconsistency residual, total-intensity component, and linear-polarization components can be adopted. The present form is selected as a simple and physically interpretable realization.}
\begin{equation}
	\mathbf{W}
	=
	\begin{bmatrix}
		\frac{1}{2} & -\frac{1}{2} & \frac{1}{2} & -\frac{1}{2} \\[1mm]
		\frac{1}{2} & \frac{1}{2} & \frac{1}{2} & \frac{1}{2} \\[1mm]
		\frac{1}{\sqrt{2}} & 0 & -\frac{1}{\sqrt{2}} & 0 \\[1mm]
		0 & \frac{1}{\sqrt{2}} & 0 & -\frac{1}{\sqrt{2}}
	\end{bmatrix},
	\label{eq:stokes_transform}
\end{equation}
which satisfies
$\mathbf{W}^{\mathrm T}\mathbf{W}=\mathbf{I}$.
Applying $\mathbf{W}$ to $\mathbf{g}_{d}$ yields
\begin{equation*}
	\mathbf{W}\mathbf{g}_{d}
	=
	\operatorname{stack}\big(
	\nabla_{d}\mathcal{C},
	\nabla_{d}\overline{\mathcal{S}}_{0},
	\frac{1}{\sqrt{2}}\nabla_{d}\mathcal{S}_{1},
	\frac{1}{\sqrt{2}}\nabla_{d}\mathcal{S}_{2}
	\big),
\end{equation*}
where $
	\overline{\mathcal{S}}_{0}
	=
	\frac{1}{2}
	\left(
	\mathcal{I}_{0^\circ}
	+
	\mathcal{I}_{45^\circ}
	+
	\mathcal{I}_{90^\circ}
	+
	\mathcal{I}_{135^\circ}
	\right)$
is the averaged estimate of the total intensity, and $
	\mathcal{C}
	=
	\frac{1}{2}
	\left[
	\left(
	\mathcal{I}_{0^\circ}
	+
	\mathcal{I}_{90^\circ}
	\right)
	-
	\left(
	\mathcal{I}_{45^\circ}
	+
	\mathcal{I}_{135^\circ}
	\right)
	\right]$
measures the inconsistency between the two estimates of the total intensity. Under the ideal linear polarization model,
$\overline{\mathcal{S}}_{0}=\mathcal{S}_{0}$ and
$\mathcal{C}=0$.

Accordingly, for each direction $d$, the transformed gradient field is given by
\begin{equation}
	\begin{aligned}
		\mathbf{W}
		\times_{\mathrm{comp}}
		\nabla_{d}\dot{\mathcal{X}}
		&=
		\nabla_{d}\mathcal{C}
		+
		\nabla_{d}\overline{\mathcal{S}}_{0}i
		+
		\frac{1}{\sqrt{2}}
		\nabla_{d}\mathcal{S}_{1}j
		+
		\frac{1}{\sqrt{2}}
		\nabla_{d}\mathcal{S}_{2}k,
	\end{aligned}
	\label{eq:stokes_gradient}
\end{equation}
where $\times_{\mathrm{comp}}$ denotes multiplication along the quaternion-component dimension. This arrangement places the total-intensity imbalance residual in the real part and embeds the three physically meaningful Stokes components into the imaginary parts. In the ideal case, the real part vanishes and the transformed gradient field becomes a pure quaternion. Moreover, the orthogonality of $\mathbf{W}$ ensures that the transformation is invertible and energy-preserving.

 After orthogonal transformation, the coupled spatial gradients of polarization intensity images are decomposed into the gradient of total-intensity $\nabla\mathcal{S}_{0}$, the gradients of polarization difference components $\nabla\mathcal{S}_{1}$, $\nabla\mathcal{S}_{2}$, and residual gradients $\nabla\mathcal{C}$. These terms exhibit distinct physical characteristics, in particular: the total-intensity gradient generally contains rich textures, geometric edges, and other scene structures; in contrast, the gradients of $\mathcal{S}_{1}$ and $\mathcal{S}_{2}$ are relatively sparse and mainly respond to changes in material properties and surface geometry \cite{S0Xie,surface}; the inconsistency residual $\mathcal{C}$ ideally vanishes and therefore tends to capture deviations caused by interpolation errors, measurement noise, and violations of the linear polarization model. These distinct characteristics indicate that identical regularization of all transformed components is inappropriate.

The transformed physical components exhibit different gradient distributions and dynamic ranges. Therefore, directly constructing the regularization weights from pointwise gradients may be unreliable, since individual gradient values are sensitive to local fluctuations and may incorrectly identify isolated noise as structural edges. To obtain a more stable estimate of the local structural activity, the gradient variation within a neighborhood is considered instead.

For notational convenience, the three physical components are denoted by $
	\mathcal{T}_{0}
	=
	\overline{\mathcal{S}}_{0},
	\mathcal{T}_{1}
	=
	\frac{1}{\sqrt{2}}\mathcal{S}_{1}$, and $
	\mathcal{T}_{2}
	=
	\frac{1}{\sqrt{2}}\mathcal{S}_{2}$.
For each direction $d$, the local mean of the gradient of the $m$th component is first computed as $
	\mu_{m,d}
	=
	\mathbf{H_s}
	\ast
	\nabla_d\mathcal{T}_m,
	\
	m\in\{0,1,2\}$,
and the corresponding local standard deviation is defined by
\begin{equation}
	\sigma_{m,d}
	=
	\sqrt{
		\mathbf{H_s}
		\ast
		\left|
		\nabla_d\mathcal{T}_m
		-
		\mu_{m,d}
		\right|^2
	},
	\qquad
	m\in\{0,1,2\}.
	\label{eq:local_energy}
\end{equation}
Here, $\ast$ denotes spatial convolution, and $\mathbf{H_s}$ is a normalized local averaging kernel. Compared with a single gradient value, $\sigma_{m,d}$ incorporates neighborhood information and is therefore less sensitive to isolated numerical fluctuations and noise. It provides a more stable measure of the local gradient variation of each physical component.

To place the local variations of different components on a common scale, a global reference is extracted from the total-intensity component:
\begin{equation}
	\sigma_{\mathrm{ref},d}
	=
	\operatorname{std}
	\big(
	\nabla_d\overline{\mathcal{S}}_0
	\big)
	+
	\epsilon,
	\label{eq:global_reference}
\end{equation}
where $\operatorname{std}(\cdot)$ denotes the standard deviation computed over all entries of the input gradient map, and $\epsilon>0$ prevents division by zero. The total-intensity component is adopted as the global reference because it aggregates information from all angular measurements and usually contains dense textures, edges, and other scene structures. Its global gradient distribution therefore provides a relatively stable structural scale for normalizing the local variations of all physical components.

Based on the ratio between the local gradient variation and the global structural reference, the adaptive weight maps are defined as
\begin{equation}
	\mathcal{P}_{m,d}
	=
	\exp
	\big[
	-
	\big(
	\frac{\sigma_{m,d}}
	{\sigma_{\mathrm{ref},d}}
	\big)^2
	\big],
	\qquad
	m\in\{0,1,2\}.
	\label{eq:adaptive_weight}
\end{equation}
The Gaussian decay function converts the normalized local variation into a weight in $(0,1]$. In regions containing pronounced physical structures, a relatively large $\sigma_{m,d}$ produces a smaller weight, thereby reducing the regularization strength and alleviating oversmoothing. In homogeneous regions, the weight remains relatively large, promoting the suppression of insignificant fluctuations.

The inconsistency-residual component $\mathcal{C}$ is treated separately because it does not correspond to an independent physical attribute and should ideally vanish under the linear polarization imaging model, which reduces physically inconsistent reconstruction errors and improves the reliability of subsequent DoLP and AoLP. Accordingly, for each direction $d$, the quaternion weight tensor is constructed as
\begin{equation}
	\dot{\Lambda}_{d}
	=
	\eta_{\mathcal C}
	+
	\mathcal{P}_{0,d}i
	+
	\mathcal{P}_{1,d}j
	+
	\mathcal{P}_{2,d}k,
	\label{eq:adaptive}
\end{equation}
where $\eta_{\mathcal C}>0$ controls the penalty imposed on the inconsistency-residual component. A relatively large value of $\eta_{\mathcal C}$ encourages the removal of nonphysical inconsistencies caused by measurement noise, interpolation errors, and model mismatch.

With the above spatially and component-wise adaptive weights, the transformed gradient components can be regularized within a unified quaternion framework. This leads to the following definition.

\begin{definition}[Adaptive Stokes-domain TV]
	Let
	$\dot{\mathcal{X}}
	\in
	\mathbb{H}^{n_{1}\times n_{2}\times n_{3}}$
	be a third-order quaternion tensor. Its adaptive Stokes-domain total variation regularizer is defined as
	\begin{equation}
	\mathcal{R}_{\mathrm{ASTV}}(\dot{\mathcal{X}})
	=
	\sum_{d\in\{h,v,t\}}
	\big\|
	\dot{\boldsymbol{\Lambda}}_d
	\odot_{\mathrm{comp}}
	(
	\mathbf{W}\times_{\mathrm{comp}}
	\nabla_d\dot{\mathcal{X}}
	)
	\big\|_{1, \mathrm{comp}},
		\label{eq:astv}
	\end{equation}
where $h$, $v$, and $t$ denote the horizontal, vertical, and tubal
directions, respectively; $\odot_{\mathrm{comp}}$ 
denotes component-wise multiplication between the quaternion components; $\Vert{} \cdot \Vert{}_{1, \mathrm{comp}}$ denotes the component-wise $\ell_1$-norm of a quaternion tensor. Specifically, for any quaternion tensor $\dot{\mathcal{T}} = \mathcal{T}_0 + \mathcal{T}_1 i + \mathcal{T}_2 j + \mathcal{T}_3 k$, its component-wise $\ell_1$-norm is defined as the sum of the $\ell_1$-norms of its four real-valued components, i.e., $\Vert{} \dot{\mathcal{T}} \Vert{}_{1, \mathrm{comp}} = \sum_{c=0}^3 \Vert{} \mathcal{T}_c \Vert{}_1$. 
\end{definition}

\subsection{Proposed Model}

Under the CPFA imaging mechanism, each sensor pixel records the intensity associated with only one color channel and one polarization orientation in a single exposure. Thus, only one of the twelve color--polarization measurements is observed at each spatial location, while the remaining eleven are missing. The central objective of CPDM is therefore to recover the complete color polarization information from the mosaicked observation $\dot{\mathcal{Y}}$. For a unified mathematical formulation, the CPFA sampling and quaternion encoding process is represented by the linear degradation operator $\mathfrak{A}(\cdot)$ as
\begin{equation}
	\begin{aligned}
		\dot{\mathcal{Y}}
		=
		\mathfrak{A}\!\big(\dot{\mathcal{X}}\big)
		:={}&
		\mathcal{M}_{0^\circ}\odot\mathcal{I}_{0^\circ}
		+
		\left(
		\mathcal{M}_{45^\circ}\odot\mathcal{I}_{45^\circ}
		\right)i \\
		&+
		\left(
		\mathcal{M}_{90^\circ}\odot\mathcal{I}_{90^\circ}
		\right)j
		+
		\left(
		\mathcal{M}_{135^\circ}\odot\mathcal{I}_{135^\circ}
		\right)k.
	\end{aligned}
	\label{eq:cpfa_observation}
\end{equation}
Here, $\mathcal{M}_{\theta}$, with
$\theta\in\{0^\circ,45^\circ,90^\circ,135^\circ\}$, denotes the binary mask tensor associated with the CPFA sampling pattern at polarization orientation $\theta$, and $\odot$ denotes the element-wise product. Thus, each masked component retains the samples observed at the corresponding orientation and sets the unobserved entries to zero. Accordingly, color polarization demosaicking can be formulated as an ill-posed inverse problem of recovering the complete quaternion tensor
$\dot{\mathcal{X}}
=
\mathcal{I}_{0^\circ}
+
\mathcal{I}_{45^\circ}i
+
\mathcal{I}_{90^\circ}j
+
\mathcal{I}_{135^\circ}k$
from the sparsely sampled observation $\dot{\mathcal{Y}}$.

\begin{figure}[h]
	\centering
	\includegraphics[width=\linewidth]{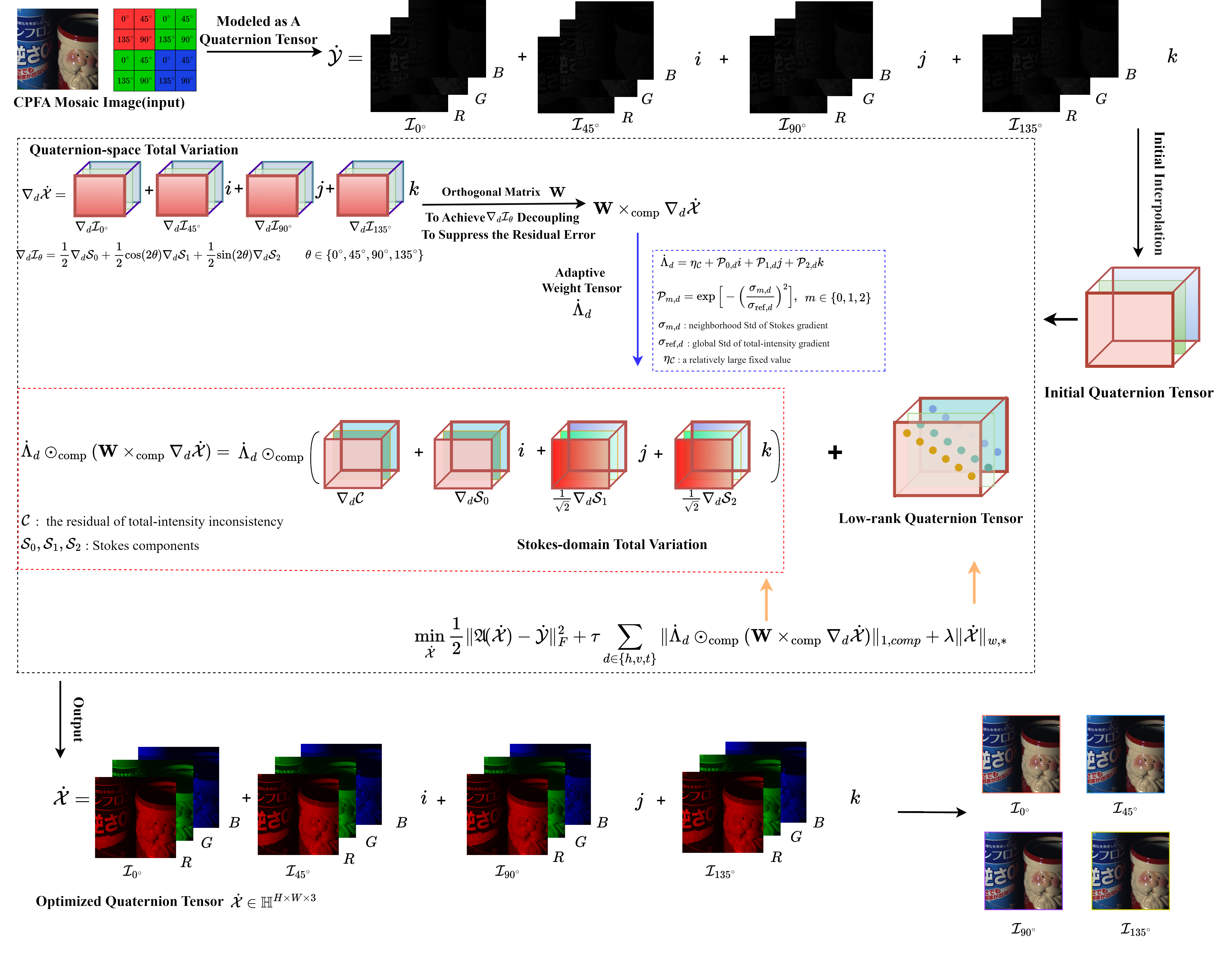}
	\caption{Overall framework of the proposed quaternion-tensor-based CPDM method.}
	\label{fig:flowchart}
\end{figure}

Motivated by the three observations established in Section \ref{sec:analysis}, a unified quaternion tensor model is proposed for CPDM task. First, the strong correlations among the four polarization channels motivate their joint representation and reconstruction within a quaternion tensor framework, rather than processing each angular image independently. Second, the rapid decay of the tensor singular values in the TQt-rank sense motivates the incorporation of a global low-rank prior to exploit their shared nonlocal structures. Third, the distinct spatial characteristics of the Stokes components motivate the use of adaptive Stokes-domain TV regularization, which preserves meaningful intensity and polarization structures while suppressing insignificant fluctuations and nonphysical inconsistencies. Based on these considerations, the proposed model is formulated as
\begin{equation}
	\begin{aligned}
		\min_{\dot{\mathcal{X}}}\quad
		&
		\frac{1}{2}
		\big\|
		\mathfrak{A}\!(\dot{\mathcal{X}})
		-
		\dot{\mathcal{Y}}
		\big\|_{F}^{2}
		+
		\lambda
		\big\|
		\dot{\mathcal{X}}
		\big\|_{w,*}+
		\tau
		\mathcal{R}_{\mathrm{ASTV}}(\dot{\mathcal{X}}).
	\end{aligned}
	\label{eq:proposed_model}
\end{equation}
The first term enforces consistency between the reconstructed tensor and the observed CPFA measurements. The QT-WNN promotes the global low-rank structure of the color polarization data, whereas the adaptive Stokes-domain TV term imposes spatially and component-wise regularization on the transformed gradients. The parameters $\lambda>0$ and $\tau>0$ balance the contributions of these two priors. The overall framework of the proposed method is illustrated in Figure \ref{fig:flowchart}.


\section{Optimization}\label{sec:4}
In this section, we present the optimization algorithm for solving the proposed model (\ref{eq:proposed_model}) and analyze its computational complexity.

\subsection{Optimization Algorithm}
The proposed model involves a nonconvex weighted low-rank regularizer, a nonsmooth adaptive Stokes-domain TV term. To decouple these terms and obtain tractable subproblems, an alternating direction method of multipliers (ADMM) framework is adopted.

Specifically, an auxiliary quaternion tensor $\dot{\mathcal{Z}}$ is introduced for the weighted low-rank term, another auxiliary quaternion tensor $\dot{\mathcal{K}}$ is introduced to separate the sampling operator in the data-fidelity term, and gradient-domain auxiliary tensors $\{\dot{\mathcal{M}}_d\}_{d\in\{h,v,t\}}$ are introduced for the adaptive Stokes-domain TV term. The corresponding variable-splitting constraints are
\begin{equation}
	\dot{\mathcal{Z}}
	=
	\dot{\mathcal{X}},
	\qquad
	\dot{\mathcal{K}}
	=
	\dot{\mathcal{X}},
	\qquad
	\dot{\mathcal{M}}_d
	=
	\mathbf{W}
	\times_{\mathrm{comp}}
	\nabla_d\dot{\mathcal{X}},
	\quad
	d\in\{h,v,t\}.
	\label{eq:splitting_constraints}
\end{equation}

%
Using the scaled dual variables
$\dot{\mathcal{U}}_{Z}$,
$\dot{\mathcal{U}}_{K}$, and
$\{\dot{\mathcal{U}}_d\}_{d\in\{h,v,t\}}$,
the scaled augmented Lagrangian, up to terms independent of the primal variables, is given by
\begin{equation}
	\begin{aligned}
		\mathcal{L}_{\rho}
		\big(
		\dot{\mathcal{X}},
		\dot{\mathcal{Z}},
		\dot{\mathcal{K}},
		\{\dot{\mathcal{M}}_d\},
		\dot{\mathcal{U}}_{Z},
		\dot{\mathcal{U}}_{K},
		\{\dot{\mathcal{U}}_d\}
		\big)
		={}&
		\frac{1}{2}
		\big\|
		\mathfrak{A}
		(
		\dot{\mathcal{K}}
		)
		-
		\dot{\mathcal{Y}}
		\big\|_F^2
		+
		\lambda
		\big\|
		\dot{\mathcal{Z}}
		\big\|_{w,*}
		\\
		&+
		\tau
		\sum_{d\in\{h,v,t\}}
		\big\|
		\dot{\Lambda}_d
		\odot_{\mathrm{comp}}
		\dot{\mathcal{M}}_d
		\big\|_{1, \mathrm{comp}}
		\\
		&+
		\frac{\rho_0}{2}
		\big\|
		\dot{\mathcal{X}}
		-
		\dot{\mathcal{Z}}
		+
		\dot{\mathcal{U}}_{Z}
		\big\|_F^2
		+
		\frac{\rho_1}{2}
		\big\|
		\dot{\mathcal{X}}
		-
		\dot{\mathcal{K}}
		+
		\dot{\mathcal{U}}_{K}
		\big\|_F^2
		\\
		&+
		\frac{\rho_2}{2}
		\sum_{d\in\{h,v,t\}}
		\big\|
		\mathbf{W}
		\times_{\mathrm{comp}}
		\nabla_d\dot{\mathcal{X}}
		-
		\dot{\mathcal{M}}_d
		+
		\dot{\mathcal{U}}_d
		\big\|_F^2,
	\end{aligned}
	\label{eq:augmented_lagrangian}
\end{equation}
where $\rho_0$, $\rho_1$, and $\rho_2$ are positive penalty parameters. At each iteration, the adaptive weight tensors $\dot{\Lambda}_d$ are evaluated from the current estimate of $\dot{\mathcal{X}}$ and then fixed while the corresponding ADMM subproblems are solved.


\subsubsection*{1) Update of the Low-Rank Auxiliary Tensor $\dot{\mathcal{Z}}$}

With the remaining variables fixed, the $\dot{\mathcal{Z}}$-subproblem at the $(k+1)$th iteration is
\begin{equation}
	\dot{\mathcal{Z}}^{k+1}
	=
	\arg\min_{\dot{\mathcal{Z}}}
	\;
	\lambda
	\big\|
	\dot{\mathcal{Z}}
	\big\|_{w,*}
	+
	\frac{\rho_0}{2}
	\big\|
	\dot{\mathcal{Z}}
	-
	(
	\dot{\mathcal{X}}^k
	+
	\dot{\mathcal{U}}_{Z}^k
	)
	\big\|_F^2.
	\label{eq:Z_subproblem}
\end{equation}
Let $
	\dot{\mathcal{B}}_Z^k
	=
	\dot{\mathcal{X}}^k
	+
	\dot{\mathcal{U}}_{Z}^k$.
The subproblem in \eqref{eq:Z_subproblem} admits a closed-form solution through the quaternion tensor singular value thresholding operator \cite{yang2024quaternion}. Specifically, compute the TQt-SVD
\begin{equation*}
	\dot{\mathcal{B}}_Z^k
	=
	\dot{\mathcal{U}}^k
	\star_{QT}
	\dot{\mathcal{D}}^k
	\star_{QT}
	\big(
	\dot{\mathcal{V}}^k
	\big)^H.
	\label{eq:TQtSVD_BZ}
\end{equation*}
For the $\ell$th tensor singular value, define the adaptive threshold $\gamma_{\ell}^k
	=
	\frac{\lambda}
	{\rho_0
		\left(
		\sigma_{\ell}
		(
		\dot{\mathcal{B}}_Z^k
		)
		+
		\epsilon_w
		\right)}$,
where $\epsilon_w>0$ prevents division by zero. Let
\begin{equation}
	\dot{\mathcal{D}}_{\gamma}^{k}
	=
	L^{-1}
	\big[
	\big(
	L
	(
	\dot{\mathcal{D}}^k
	)
	-
	\boldsymbol{\Gamma}^k
	\big)_{+}
	\big],
	\label{eq:weighted_shrunk_singular_tensor}
\end{equation}
where $\boldsymbol{\Gamma}^k$ is the diagonal threshold tensor formed from
$\{\gamma_{\ell}^k\}$, $L(\cdot)$ denotes the transform employed in the TQt-SVD, $L^{-1}(\cdot)$ is its inverse, and $(\cdot)_{+}$ denotes the positive-part operator. The update of $\dot{\mathcal{Z}}$ is then
\begin{equation}
	\dot{\mathcal{Z}}^{k+1}
	=
	\dot{\mathcal{U}}^k
	\star_{QT}
	\dot{\mathcal{D}}_{\gamma}^{k}
	\star_{QT}
	(
	\dot{\mathcal{V}}^k
	)^H.
	\label{eq:sub_Z}
\end{equation}

\subsubsection*{2) Update of the Data-Fidelity Auxiliary Tensor $\dot{\mathcal{K}}$}

With the remaining variables fixed, the $\dot{\mathcal{K}}$-subproblem is
\begin{equation}
	\dot{\mathcal{K}}^{k+1}
	=
	\arg\min_{\dot{\mathcal{K}}}
	\;
	\frac{1}{2}
	\big\|
	\mathfrak{A}
	(
	\dot{\mathcal{K}}
	)
	-
	\dot{\mathcal{Y}}
	\big\|_F^2
	+
	\frac{\rho_1}{2}
	\big\|
	\dot{\mathcal{K}}
	-
(
	\dot{\mathcal{X}}^k
	+
	\dot{\mathcal{U}}_{K}^k
	)
	\big\|_F^2.
	\label{eq:K_subproblem}
\end{equation}
Let $\dot{\mathcal{B}}_K^k  
	=
	\dot{\mathcal{X}}^k
	+
	\dot{\mathcal{U}}_{K}^k$.
The optimality condition of \eqref{eq:K_subproblem} gives
\begin{equation}
	\mathfrak{A}^{\triangleright}
	\big(
	\mathfrak{A}
	(
	\dot{\mathcal{K}}^{k+1}
	)
	\big)
	+
	\rho_1\dot{\mathcal{K}}^{k+1}
	=
	\mathfrak{A}^{\triangleright}
	(
	\dot{\mathcal{Y}}
	)
	+
	\rho_1\dot{\mathcal{B}}_K^k.
	\label{eq:K_normal_equation}
\end{equation}
where $\mathfrak{A}^{\triangleright}$ denotes the adjoint of $\mathfrak{A}$. Since the CPFA sampling operator is a binary masking operator, $\mathfrak{A}^{\triangleright}\mathfrak{A}$ is diagonal and satisfies
$\mathfrak{A}^{\triangleright}\mathfrak{A}=\mathfrak{A}$.
Therefore, the subproblem can be solved element-wise as
\begin{equation}
	\dot{\mathcal{K}}^{k+1}
	=
	\dot{\mathcal{B}}_K^k
	+
	\frac{1}{1+\rho_1}
	\mathfrak{A}^{\triangleright}
	\big[
	\dot{\mathcal{Y}}
	-
	\mathfrak{A}
	(
	\dot{\mathcal{B}}_K^k
	)
	\big].
	\label{eq:sub_K}
\end{equation}
Thus, the observed entries are jointly determined by the measurements and the current estimate, whereas the unobserved entries remain equal to those of $\dot{\mathcal{B}}_K^k$.

\subsubsection*{3) Update of the Gradient-Domain Auxiliary Tensors $\dot{\mathcal{M}}_d$}

For each direction $d\in\{h,v,t\}$, the corresponding subproblem is
\begin{equation}
	\begin{aligned}
		\dot{\mathcal{M}}_d^{k+1}
		=
		\arg\min_{\dot{\mathcal{M}}_d}
		\quad
		&
		\tau
		\big\|
		\dot{\Lambda}_d^k
		\odot_{\mathrm{comp}}
		\dot{\mathcal{M}}_d
		\big\|_{1, \mathrm{comp}}
		\\
		&+
		\frac{\rho_2}{2}
		\big\|
		\dot{\mathcal{M}}_d
		-
		(
		\mathbf{W}
		\times_{\mathrm{comp}}
		\nabla_d\dot{\mathcal{X}}^k
		+
		\dot{\mathcal{U}}_d^k
		)
		\big\|_F^2.
	\end{aligned}
	\label{eq:M_subproblem}
\end{equation}
Define $
	\dot{\mathcal{G}}_d^k
	=
	\mathbf{W}
	\times_{\mathrm{comp}}
	\nabla_d\dot{\mathcal{X}}^k
	+
	\dot{\mathcal{U}}_d^k$.
The subproblem in \eqref{eq:M_subproblem} is the proximal mapping of a component-wise weighted $\ell_1$ norm and therefore admits the closed-form solution
\begin{equation}
	\dot{\mathcal{M}}_d^{k+1}
	=
	\operatorname{Shrink}_{\frac{\tau}{\rho_2}
		\dot{\Lambda}_d^k}^{\mathrm{comp}}
	(
	\dot{\mathcal{G}}_d^k
	),
	\qquad
	d\in\{h,v,t\},
	\label{eq:sub_M}
\end{equation}
where
$\operatorname{Shrink}^{\mathrm{comp}}(\cdot)$
denotes component-wise soft thresholding. More precisely, for
\begin{equation*}
	\dot{\mathcal{G}}
	=
	\mathcal{G}_0
	+
	\mathcal{G}_1i
	+
	\mathcal{G}_2j
	+
	\mathcal{G}_3k \ \text{and}\
	\dot{\Theta}
	=
	\Theta_0
	+
	\Theta_1i
	+
	\Theta_2j
	+
	\Theta_3k,
\end{equation*}
the operator is defined by
\begin{equation}
	\operatorname{Shrink}_{\dot{\Theta}}^{\mathrm{comp}}
	(
	\dot{\mathcal{G}}
	)
	=
	\sum_{\ell=0}^{3}
	\operatorname{sign}
	(
	\mathcal{G}_{\ell}
	)
	\odot
	\max
	\left(
	|
	\mathcal{G}_{\ell}
	|
	-
	\Theta_{\ell},
	0
	\right)e_{\ell},\quad \ell\in\{0,1,2,3\},
	\label{eq:componentwise_shrinkage}
\end{equation}
where
$e_0=1$, $e_1=i$, $e_2=j$, and $e_3=k$.

\subsubsection*{4) Update of the Primal Variable $\dot{\mathcal{X}}$}

With the remaining variables fixed, the update of $\dot{\mathcal{X}}$ is obtained by solving
\begin{equation}
	\begin{aligned}
		\dot{\mathcal{X}}^{k+1}
		=
		\arg\min
		_{\dot{\mathcal{X}}}
		\quad
		&
		\frac{\rho_0}{2}
		\big\|
		\dot{\mathcal{X}}
		-
		\dot{\mathcal{Z}}^{k+1}
		+
		\dot{\mathcal{U}}_{Z}^k
		\big\|_F^2+
		\frac{\rho_1}{2}
		\big\|
		\dot{\mathcal{X}}
		-
		\dot{\mathcal{K}}^{k+1}
		+
		\dot{\mathcal{U}}_{K}^k
		\big\|_F^2
		\\
		&+
		\frac{\rho_2}{2}
		\sum_{d\in\{h,v,t\}}
		\big\|
		\mathbf{W}
		\times_{\mathrm{comp}}
		\nabla_d\dot{\mathcal{X}}
		-
		\dot{\mathcal{M}}_d^{k+1}
		+
		\dot{\mathcal{U}}_d^k
		\big\|_F^2.
	\end{aligned}
	\label{eq:X_constrained_subproblem}
\end{equation}

The quadratic part of \eqref{eq:X_constrained_subproblem} is minimized, yielding an solution
$\dot{\mathcal{X}}^{k+1}$.
By using the orthogonality relation
$\mathbf{W}^{T}\mathbf{W}=\mathbf{I}$,
the corresponding normal equation is
\begin{equation}\label{eq:X_normal_equation}
	(\rho_0+\rho_1)\dot{\mathcal{X}}^{k+1}
	+
	\rho_2
	\sum_{d\in\{h,v,t\}}
	\nabla_d^{\triangleright}\nabla_d
	\dot{\mathcal{X}}^{k+1}
	=
	\dot{\mathcal{B}}_X^k.
\end{equation}
where
\begin{equation*}
		\dot{\mathcal{B}}_X^k
		={}
		\rho_0
		(
		\dot{\mathcal{Z}}^{k+1}
		-
		\dot{\mathcal{U}}_{Z}^k
		)
		+
		\rho_1
		(
		\dot{\mathcal{K}}^{k+1}
		-
		\dot{\mathcal{U}}_{K}^k
		)+
		\rho_2
		\sum_{d\in\{h,v,t\}}
		\nabla_d^{\triangleright}
		\big[
		\mathbf{W}^{T}
		\times_{\mathrm{comp}}
		(
		\dot{\mathcal{M}}_d^{k+1}
		-
		\dot{\mathcal{U}}_d^k
		)
		\big].
\end{equation*}
Here, $\nabla_d^{\triangleright}$ denotes the adjoint of the directional finite-difference operator $\nabla_d$.


Under periodic boundary conditions, the operators
$\nabla_d^{\triangleright}\nabla_d$ have block-circulant
structures and can be diagonalized by the discrete Fourier
transform. Let $\widehat{\mathcal{D}}_d$ denote the frequency
response of $\nabla_d$, and define
\begin{equation}
	\widehat{\mathcal{D}}
	=
	(\rho_0+\rho_1)
	+
	\rho_2
	\sum_{d\in\{h,v,t\}}
	\left|
	\widehat{\mathcal{D}}_d
	\right|^2.
	\label{eq:frequency_denominator}
\end{equation}
Since the finite-difference operators have real-valued
coefficients and act identically on the four quaternion
components, the resulting linear system can be diagonalized
component-wise in the Fourier domain. Moreover,
$\widehat{\mathcal{D}}$ is real-valued and therefore commutes
with arbitrary quaternion entries. Consequently, the
final solution is given by
\begin{equation}
	\dot{\mathcal{X}}^{k+1}
	=
	\mathcal{F}_{\mathrm{comp}}^{-1}
	\left[
	\mathcal{F}_{\mathrm{comp}}
	\left(
	\dot{\mathcal{B}}_X^k
	\right)
	\oslash
	\widehat{\mathcal{D}}
	\right],
	\label{eq:X_unconstrained_solution}
\end{equation}
where $\mathcal{F}_{\mathrm{comp}}$ and
$\mathcal{F}_{\mathrm{comp}}^{-1}$ denote the three-dimensional
Fourier transform and its inverse applied separately to the four components of a quaternion tensor, respectively; \(\oslash\) denotes element-wise division.

\subsubsection*{5) Update of the Scaled Dual Variables}

The scaled dual variables associated with the splitting constraints are updated as
\begin{subequations}
	\label{eq:update_all_multipliers}
	\begin{align}
		\dot{\mathcal{U}}_{Z}^{k+1}
		={}&
		\dot{\mathcal{U}}_{Z}^k
		+
		\dot{\mathcal{X}}^{k+1}
		-
		\dot{\mathcal{Z}}^{k+1},
		\label{eq:update_U0}
		\\
		\dot{\mathcal{U}}_{K}^{k+1}
		={}&
		\dot{\mathcal{U}}_{K}^k
		+
		\dot{\mathcal{X}}^{k+1}
		-
		\dot{\mathcal{K}}^{k+1},
		\label{eq:update_U1}
		\\
		\dot{\mathcal{U}}_d^{k+1}
		={}&
		\dot{\mathcal{U}}_d^k
		+
		\mathbf{W}
		\times_{\mathrm{comp}}
		\nabla_d\dot{\mathcal{X}}^{k+1}
		-
		\dot{\mathcal{M}}_d^{k+1},\
		d\in\{h,v,t\}.
		\label{eq:update_Ud}
	\end{align}
\end{subequations}

Algorithm~\ref{alg:jcpd_admm} summarizes the overall optimization procedure of the proposed method.

\begin{algorithm}[t]
\caption{Optimization Procedure for the Proposed Quaternion-Tensor-Based CPDM Model}
	\label{alg:jcpd_admm}
	\begin{algorithmic}[1]
		\renewcommand{\algorithmicrequire}{\textbf{Input:}}
		\renewcommand{\algorithmicensure}{\textbf{Output:}}
		
		\REQUIRE Observed quaternion tensor $\dot{\mathcal{Y}}\in \mathbb{H}^{I_1\times I_2\times I_3}$;
		invertible quaternion matrix $\dot{\mathbf{Q}}_3\in \mathbb{H}^{I_3\times I_3}$;
		 sampling operator $\mathfrak{A}$; transformation matrix $\mathbf{W}$;
		 normalized 5×5 local averaging kernel \(\mathbf{H_s}\);
		 a relatively large positive value \(\eta_{\mathcal C}\);
		  parameters $\lambda$, $\tau$, $\rho_0$, $\rho_1$, $\rho_2$, $\eta_{\mathcal C}$, $\epsilon_w$,$\epsilon$; maximum iteration number $K_{\max}$;  tol (tolerance $\mathrm{tol})$.
		\ENSURE Reconstructed quaternion tensor $\dot{\mathcal{X}}\in \mathbb{H}^{I_1\times I_2\times I_3}$.
		
		\STATE Obtain an initial estimate $\dot{\mathcal{X}}_{\mathrm{init}}$ using a conventional interpolation method.
		\STATE Set
		$\dot{\mathcal{X}}^{0}
		=\dot{\mathcal{X}}_{\mathrm{init}}$.
		\STATE Initialize
		$\dot{\mathcal{Z}}^{0}
		=\dot{\mathcal{X}}^{0}$, $\dot{\mathcal{K}}^{0}=\dot{\mathcal{X}}^{0}$,
		$\dot{\mathcal{M}}_d^{0}
		=
		\mathbf{W}
		\times_{\mathrm{comp}}
		\nabla_d\dot{\mathcal{X}}^{0}$,
		$\dot{\mathcal{U}}_d^{0}
		=\mathbf{0}$
		for all $d\in\{h,v,t\}$, and $\dot{\mathcal{U}}_{Z}^{0}
		=
		\dot{\mathcal{U}}_{K}^{0}
		=
		\mathbf{0}$
		\FOR{$k=0,1,\ldots,K_{\max}-1$}
		
		\STATE Update the weighted low-rank auxiliary tensor
		$\dot{\mathcal{Z}}^{k+1}$
		via \eqref{eq:sub_Z}.
		
		\STATE Update the data-fidelity auxiliary tensor
		$\dot{\mathcal{K}}^{k+1}$
		via \eqref{eq:sub_K}.
		
		\STATE Compute the adaptive weight tensors
		$\{\dot{\Lambda}_d^k\}_{d\in\{h,v,t\}}$
		from $\dot{\mathcal{X}}^k$.
		
		\STATE Update
		$\{\dot{\mathcal{M}}_d^{k+1}\}_{d\in\{h,v,t\}}$
		via \eqref{eq:sub_M}.
		
		\STATE Compute the primal variable
		$\dot{\mathcal{X}}^{k+1}$
		via \eqref{eq:X_unconstrained_solution}.
		
		
		\STATE Update the scaled dual variables via
		\eqref{eq:update_U0}--\eqref{eq:update_Ud}.
		
		\STATE Compute
		\[
		\mathrm{RelChg}^{k+1}
		=
		\frac{
			\|
			\dot{\mathcal{X}}^{k+1}
			-
			\dot{\mathcal{X}}^{k}
			\|_F
		}{
			\max
			\big\{
			\|
			\dot{\mathcal{X}}^{k}
			\|_F,
			\epsilon
			\big\}
		}.
		\]
		
		\IF{$\mathrm{RelChg}^{k+1}<\mathrm{tol}$}
		\STATE \textbf{break}
		\ENDIF
		
		\ENDFOR
		
		\STATE Set
		$\dot{\mathcal{X}}
		=
		\dot{\mathcal{X}}^{k+1}$.
		
		\RETURN $\dot{\mathcal{X}}$.
	\end{algorithmic}
\end{algorithm}

\subsection{Computational Complexity Analysis}
\label{sec:complexity}

Let the spatial resolution of each color polarization image be
$H\times W$, and let $C$ denote the number of color channels.
In the present application, $C=3$ is fixed. The quaternion tensor
to be reconstructed therefore has size $H\times W\times C$.
Let $K$ denote the total number of iterations.

The initialization step interpolates the CPFA mosaic to construct
an initial quaternion tensor, requiring
$\mathcal{O}(CHW)$ operations. At each iteration, the principal
computational costs arise from the following updates.

The update of $\dot{\mathcal{Z}}$ requires a TQt-SVD. After applying
the transform along the third mode, quaternion matrix singular value
decompositions are performed on $C$ frontal slices of size
$H\times W$. Therefore, the corresponding computational complexity is $
	\mathcal{O}
	\left(
	CHW\min\{H,W\}
	\right)$.
The transform and inverse transform along the third mode require only
$\mathcal{O}(CHW)$ operations when $C$ is fixed and are therefore
absorbed into the lower-order terms.

The update of $\dot{\mathcal{K}}$ consists of element-wise operations
associated with the binary sampling masks and has complexity $\mathcal{O}(CHW)$.
The computation of the adaptive weight tensors involves finite
differences, local convolutions, and element-wise operations. Since
the number of gradient directions, physical components, and the size
of the local averaging kernel are fixed, its complexity is also $
\mathcal{O}(CHW)$.
Similarly, the updates of
$\{\dot{\mathcal{M}}_d\}_{d\in\{h,v,t\}}$
through component-wise soft thresholding require
$\mathcal{O}(CHW)$ operations.

For the update of $\dot{\mathcal{X}}$, the right-hand side of
\eqref{eq:X_normal_equation} is first constructed using finite
differences and component transformations, with complexity
$\mathcal{O}(CHW)$. The resulting linear system is then solved by
applying the three-dimensional discrete Fourier transform separately
to the four real components of the quaternion tensor. Consequently,
the frequency-domain solution requires$
\mathcal{O}
	\left(
	CHW\log(CHW)
	\right)$
operations, where the constant factor associated with the four
quaternion components is omitted. 


Finally, the updates of the scaled dual variables and the evaluation
of the stopping criterion involve only finite differences,
tensor additions, and norm calculations, all of which require
$\mathcal{O}(CHW)$ operations. Hence, the overall computational
complexity per iteration is $
	\mathcal{O}
	\left(
	CHW\min\{H,W\}
	+
	CHW\log(CHW)
	+
	CHW
	\right)$.
For typical high-resolution images, the TQt-SVD constitutes the
dominant computational cost. Therefore, the per-iteration complexity
can be simplified as $
	\mathcal{O}
	\left(
	CHW\min\{H,W\}
	\right)$.
Including the initialization and $K$ iterations, the total
computational complexity is $
	\mathcal{O}
	\left(
	CHW
	+
	KCHW
	\left[
	\min\{H,W\}
	+
	\log(CHW)
	\right]
	\right)$.
Since $C=3$ is fixed and the TQt-SVD dominates the remaining
operations, the total complexity may be further expressed as $
	\mathcal{O}
	\left(
	KHW\min\{H,W\}
	\right)$.

\section{Experiments and Analysis}\label{sec:5}

In this section, the performance of the proposed method is evaluated using the peak signal-to-noise ratio (PSNR) \cite{chen2015denoising}, structural similarity index measure (SSIM) \cite{jordan2012predicting}, and root-mean-square error (RMSE) of the AoLP \cite{yi2024demosaicking}, together with visual comparisons. The effectiveness and advantages of the proposed method are demonstrated through comparisons with several state-of-the-art approaches.

The compared methods include:
\begin{enumerate}[1)]
	\item Interpolation-based methods: EARI \cite{morimatsu2020monochrome} and PCDP \cite{wu2021polarization};
	\item Optimization-based methods: LMMSE \cite{dumoulin2022impact}, SR-JCPD \cite{wen2021sparse}, and NLCSR \cite{luo2024learning};
	\item Deep learning-based methods: TCPDNet \cite{nguyen2022two} and PUGDiff \cite{Li_Luo_Zhang_Yang_2026}.
\end{enumerate}

\begin{figure}[!h]
	\centering
	
	\includegraphics[width=0.8\textwidth]
	{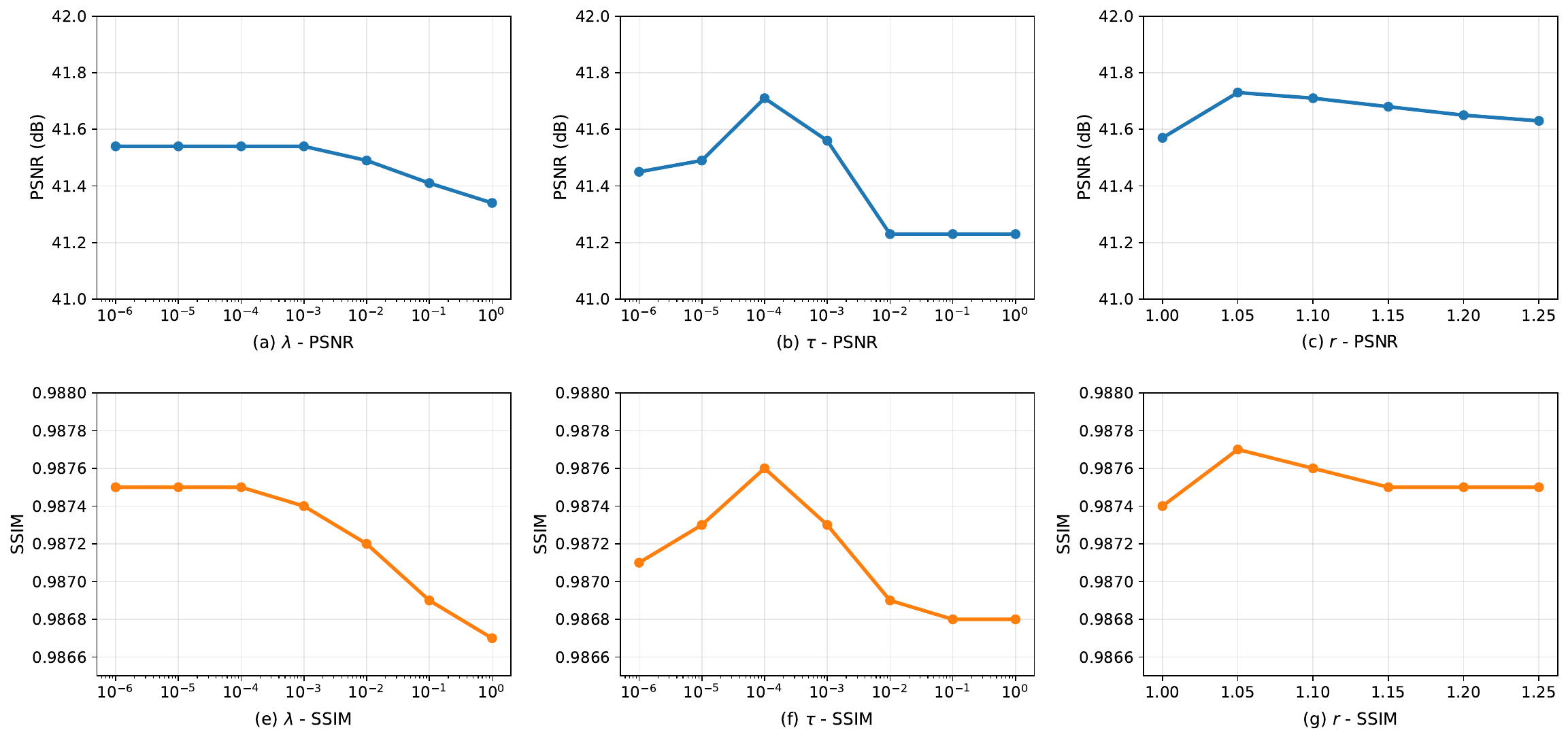}
	
	\vspace{-0.5em}
	\centerline{(i) Wen dataset}
	
	\vspace{0.6em}
	
	\includegraphics[width=0.8\textwidth]
	{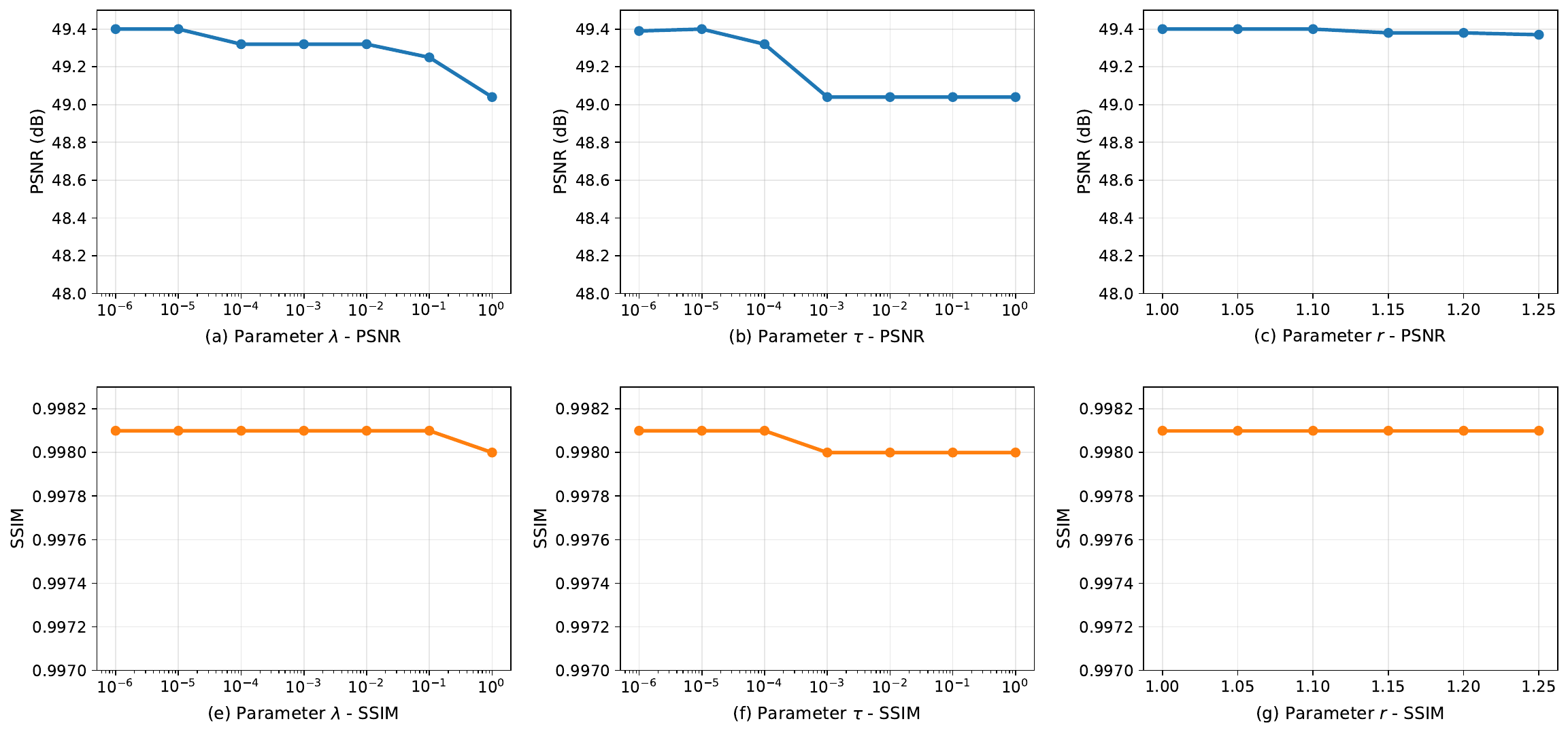}
	
	\vspace{-0.5em}
	\centerline{(ii) Guo dataset}
	
	\caption{Parameter sensitivity analysis on the Wen and Guo datasets.}
	\label{fig:param_sensitivity}
\end{figure}

\begin{figure}[h]
	\centering
	\includegraphics[width=0.8\textwidth]{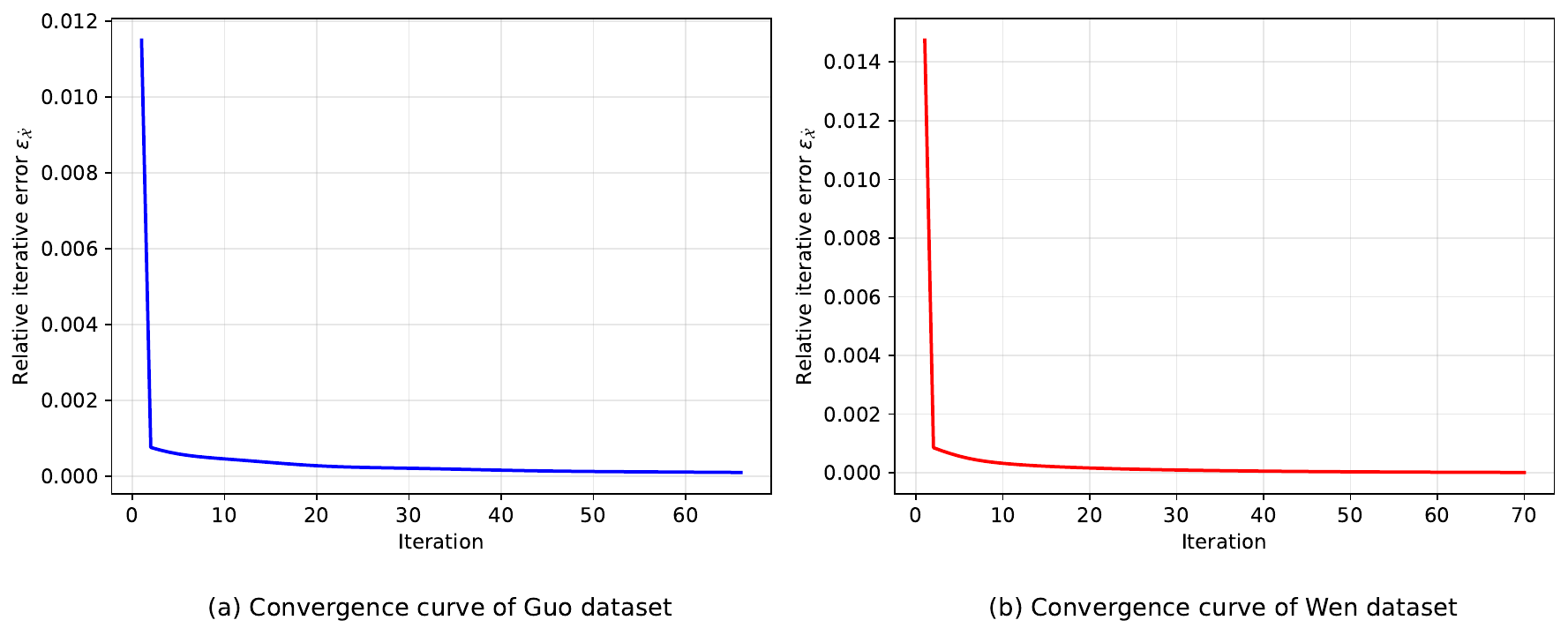}
	\caption{Convergence behavior of the proposed algorithm on the Wen and Guo datasets.}
	\label{fig:convergence}
\end{figure}

The experiments are conducted on two publicly available datasets, namely, the datasets provided by Wen et al. \cite{Wen_Zheng_Lu_Zhao_2019}  and Guo et al. \cite{zhou2026pidsr}. Both datasets contain complete high-resolution color polarization images acquired at four polarization orientations. To investigate the impact of hyperparameters on the reconstruction performance of the proposed algorithm, a comprehensive parameter sensitivity analysis is conducted regarding the regularization parameters $\lambda$ and $\tau$, the penalty parameter step size $r$ across the utilized datasets. As illustrated in Figure \ref{fig:param_sensitivity}, the proposed algorithm exhibits robust performance across a wide range of the investigated parameters. By comprehensively balancing the quantitative outcomes from both datasets, the algorithm achieves the optimal and consistent reconstruction quality when setting $\lambda = 0.0001$, $\tau = 0.001$ and $r = 1.05$. Consequently, this specific hyperparameter configuration is fixed as the global optimal deployment for all subsequent comparative experiments. In addition, the convergence behavior of the proposed algorithm on the Wen and Guo datasets is illustrated in Figure \ref{fig:convergence}. The horizontal axis represents the iteration index $k$, while the vertical axis represents the relative change between two consecutive iterates, defined as
\begin{equation}
	\varepsilon_{\dot{\mathcal{X}}}^{(k)}
	=
	\frac{
		\|
		\dot{\mathcal{X}}^{(k)}
		-
		\dot{\mathcal{X}}^{(k-1)}
		\|_F
	}{
		\|
		\dot{\mathcal{X}}^{(k-1)}
		\|_F
	}.
\end{equation}
As shown in the figure \ref{fig:convergence}, the relative changes decrease rapidly and gradually stabilize at small values, demonstrating the fast numerical convergence of the proposed algorithm.

\begin{figure*}[!tbp]
	\centering
	\captionsetup{justification=raggedright, singlelinecheck=false}
	\includegraphics[width=0.9\linewidth]{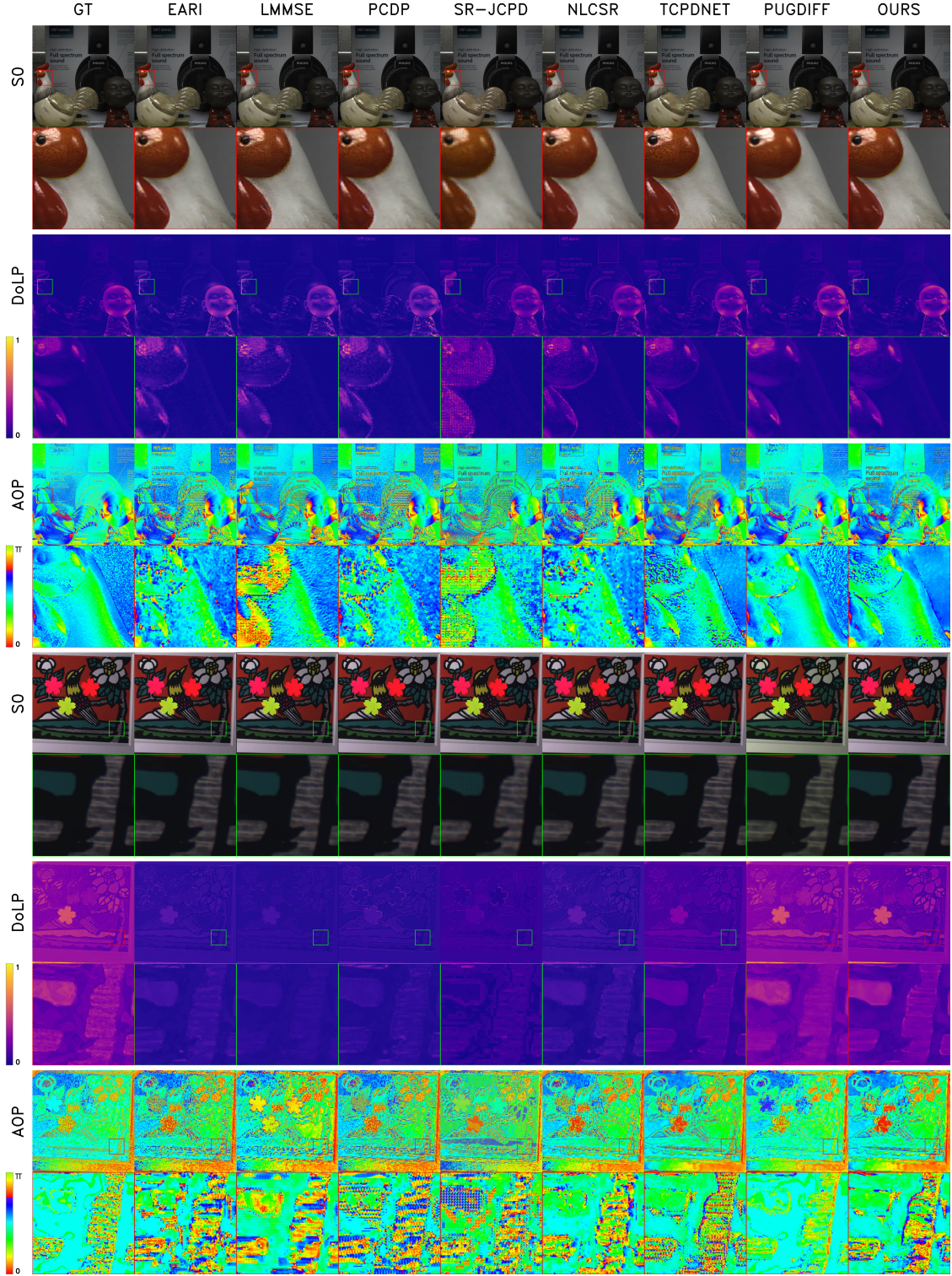}
	\caption{Visual comparison of the reconstructed $\mathcal{S}_0$, DoLP and AoLP maps on representative scenes. From left to right: Ground Truth (GT), EARI, LMMSE, PCDP, SR-JCPD, NLCSR, TCPDNet, PUGDiff, and the proposed method (Ours).}
	\label{fig:visual}
\end{figure*}

Tables~\ref{tab:wen_comparison} and~\ref{tab:guo_comparison} report the average PSNR and SSIM values for the reconstructed $I_{0^\circ}$, $I_{45^\circ}$, $I_{90^\circ}$, $I_{135^\circ}$, $S_0$, and DoLP images, together with the average AoLP error on the Wen and Guo datasets, respectively. Higher PSNR and SSIM values and a lower AoLP error indicate better performance in the CPDM task. The best results are highlighted in bold, while the second-best results are underlined. At the same time, we perform a qualitative analysis to visually demonstrate the superiority of the proposed algorithm. Figure \ref{fig:visual} illustrates the visual comparison of the demosaicked $\mathcal{S}_0$, AoLP, and DoLP maps on representative scenes from the utilized datasets. Magnified patches of key regions are highlighted to allow careful observation of high frequency details and structural fidelity. The overall results demonstrate that the proposed method achieves clear advantages over existing approaches in both structural fidelity and physical consistency.

\begin{table*}[h]
	\centering
	\caption{Quantitative comparison on the Wen dataset (PSNR: dB; SSIM: unitless; AoLP error: degrees).}
	\label{tab:wen_comparison}
	\renewcommand{\arraystretch}{1.35}
	\resizebox{\linewidth}{!}{
		\begin{tabular}{c c ccccccc ccccccc c}
			\toprule
			\multirow{2}{*}{\textbf{Category}} 
			& \multirow{2}{*}{\textbf{Method}} 
			& \multicolumn{7}{c}{\textbf{PSNR}} 
			& \multicolumn{7}{c}{\textbf{SSIM}} 
			& \multirow{2}{*}{\textbf{AoLP error}} \\
			\cmidrule(lr){3-9} 
			\cmidrule(lr){10-16}
			& 
			& $I_{\mathrm{mean}}$ 
			& $I_{0^\circ}$ 
			& $I_{45^\circ}$ 
			& $I_{90^\circ}$ 
			& $I_{135^\circ}$ 
			& $S_{0}$ 
			& DoLP 
			& $I_{\mathrm{mean}}$ 
			& $I_{0^\circ}$ 
			& $I_{45^\circ}$ 
			& $I_{90^\circ}$ 
			& $I_{135^\circ}$ 
			& $S_{0}$ 
			& DoLP 
			& \\
			\midrule
			
			\multirow{2}{*}{\small Interpolation} 
			& PCDP
			& 39.55 & 39.17 & 39.66 & 39.84 & 39.54 & 40.36 & 31.96
			& 0.9886 & 0.9874 & 0.9882 & 0.9903 & 0.9883 & 0.9906 & 0.8853
			& 15.30 \\
			
			& EARI
			& 39.57 & 38.87 & 39.57 & 40.09 & 39.76 & 41.99 & 32.94
			& \underline{0.9913} & \underline{0.9900} & \underline{0.9912}
			& \underline{0.9924} & \underline{0.9916} & \underline{0.9922}
			& \underline{0.9131}
			& \underline{14.53} \\
			\midrule
			
			\multirow{3}{*}{\small Optimization} 
			& LMMSE
			& 39.73 & 39.11 & 39.89 & 40.13 & 39.79 & 40.75 & 33.17
			& 0.9718 & 0.9707 & 0.9719 & 0.9730 & 0.9717 & 0.9733 & 0.8340
			& 17.75 \\
			
			& SR-JCPD
			& 37.77 & 37.42 & 38.08 & 37.81 & 37.78 & 39.28 & 30.07
			& 0.9851 & 0.9839 & 0.9852 & 0.9860 & 0.9852 & 0.9877 & 0.7907
			& 22.34 \\
			
			& NLCSR
			& 41.36 & 41.25 & 41.43 & 41.41 & 41.37 & 36.08 & 36.05
			& 0.9791 & 0.9783 & 0.9796 & 0.9795 & 0.9791 & 0.9706 & 0.8960
			& 14.70 \\
			\midrule
			
			\multirow{2}{*}{\small Deep Learning} 
			& TCPDNet
			& \underline{42.99} & \textbf{42.80} & 42.97
			& \textbf{43.34} & 42.86 & \textbf{44.48} & \textbf{37.91}
			& 0.9867 & 0.9858 & 0.9868 & 0.9878 & 0.9865 & 0.9900
			& \underline{0.9053}
			& 22.65 \\
			
			& PUGDiff
			& 42.92 & 42.59 & \underline{43.10} & 42.98
			& \textbf{43.00} & 38.08 & \underline{37.18}
			& 0.9868 & 0.9859 & 0.9873 & 0.9871 & 0.9868 & 0.9838 & 0.8891
			& 15.66 \\
			\midrule
			
			& \textbf{Ours}
			& \textbf{43.00} & \underline{42.72} & \textbf{43.16}
			& \underline{43.19} & \underline{42.95} & \underline{42.71} & 36.46
			& \textbf{0.9938} & \textbf{0.9934} & \textbf{0.9938}
			& \textbf{0.9945} & \textbf{0.9936} & \textbf{0.9950}
			& \textbf{0.9291}
			& \textbf{13.45} \\
			\bottomrule
		\end{tabular}
	}
\end{table*}

\begin{table*}[h]
	\centering
	\caption{Quantitative comparison on the Guo dataset (PSNR: dB; SSIM: unitless; AoLP error: degrees).}
	\label{tab:guo_comparison}
	\renewcommand{\arraystretch}{1.35}
	\resizebox{\linewidth}{!}{
		\begin{tabular}{c c ccccccc ccccccc c}
			\toprule
			\multirow{2}{*}{\textbf{Category}} & \multirow{2}{*}{\textbf{Method}} & \multicolumn{7}{c}{\textbf{PSNR}} & \multicolumn{7}{c}{\textbf{SSIM}} & \multirow{2}{*}{\textbf{AoLP error}} \\
			\cmidrule(lr){3-9} \cmidrule(lr){10-16}
			& & $I_{mean}$ & $I_{0^\circ}$ & $I_{45^\circ}$ & $I_{90^\circ}$ & $I_{135^\circ}$ & $S_{0}$ & DoLP & $I_{mean}$ & $I_{0^\circ}$ & $I_{45^\circ}$ & $I_{90^\circ}$ & $I_{135^\circ}$ & $S_{0}$ & DoLP & \\
			\midrule
			\multirow{2}{*}{\small Interpolation} 
			& PCDP     & 38.48 & 38.11 & 38.37 & 38.87 & 38.56 & 40.27 & 29.31 & 0.9762 & 0.9748 & 0.9759 & 0.9774 & 0.9765 & 0.9839 & 0.8488 & 12.50 \\
			& EARI     & 40.56 & 40.14 & 40.35 & 41.08 & 40.67 & 43.02 & 33.98 & 0.9847 & 0.9837 & 0.9839 & 0.9861 & 0.9851 & \underline{0.9903} & 0.8927 & \underline{10.45} \\
			\midrule
			\multirow{3}{*}{\small Optimization} 
			& LMMSE    & 39.08 & 39.07 & 38.88 & 38.93 & 39.44 & 40.51 & 30.30 & 0.9580 & 0.9582 & 0.9568 & 0.9577 & 0.9506 & 0.9631 & 0.8415 & 12.47 \\
			& SR-JCPD     & 34.63 & 34.38 & 34.80 & 34.68 & 34.68 & 36.21 & 27.21 & 0.9405 & 0.9394 & 0.9421 & 0.9390 & 0.9417 & 0.9547 & 0.6915 & 18.68 \\
			& NLCSR    & 39.60 & 39.18 & 39.50 & 39.99 & 39.75 & 35.10 & 32.90 & 0.9781 & 0.9770 & 0.9775 & 0.9792 & 0.9788 & 0.9677 & 
			0.8976
			 & 10.48 \\
			\midrule
			\multirow{2}{*}{\small Deep Learning} 
			& PUGDiff     & 41.02 & 40.63 & \underline{40.88} & 41.39 & \underline{41.17} & 36.36 & 33.63 & 0.9818 & 0.9810 & 0.9816 & 0.9822 & 0.9823 & 0.9761 & 0.8556 & 14.57 \\
			& TCPDNet  & \underline{41.16} & \underline{40.99} & 40.62 & \underline{41.90} & 41.14 & \underline{43.15} & \textbf{35.48} & \underline{0.9855} & \underline{0.9850} & \underline{0.9841} & \underline{0.9870} & \underline{0.9861} & 0.9902 & \underline{0.9012} & 13.26 \\
			\midrule
			\textbf{} & \textbf{Ours}     & \textbf{41.68} & \textbf{41.26} & \textbf{41.44} & \textbf{42.11} & \textbf{41.91} & \textbf{43.56} & \underline{34.99} & \textbf{0.9881} & \textbf{0.9873} & \textbf{0.9872} & \textbf{0.9890} & \textbf{0.9890} & \textbf{0.9919} & \textbf{0.9134} & \textbf{8.83} \\
			\bottomrule
		\end{tabular}
	}
\end{table*}

As shown in tables~\ref{tab:wen_comparison} and~\ref{tab:guo_comparison}, PCDP and EARI exploit channel-difference priors combined with interpolation strategies and provide relatively stable reconstruction quality in the intensity domain. The AoLP error remains acceptable, while the overall PSNR and SSIM values are marginally inferior. As illustrated in Figure \ref{fig:visual},  the reconstructed DoLP maps for PCDP and EARI exhibit a significantly darker overall appearance than the ground truth, indicating a severe underestimation of the degree of polarization. Furthermore, fine polarization structures—such as the folds of the chick's neck—are blurred or completely lost, while their corresponding AoLP maps display disordered color patterns and distorted contours near structural boundaries. All the above reconstruction phenomena collectively demonstrate that conventional polarization interpolation methods fail to fully utilize the physical correlation of multi-polarization channels, which triggers compromised image fidelity, obvious interpolation artifacts, inaccurate DoLP estimation, lost high-frequency texture details and severe distortion of fine polarization structures and edge contours.
 
LMMSE relies on statistical information estimated from training data. Its performance may therefore be sensitive to discrepancies between the training and test data distributions.  In addition, its linear estimation mechanism has limited capability to preserve complex spatial structures, which may account for its relatively low SSIM values across both datasets. Visually, LMMSE produces globally dim DoLP maps with reduced polarization contrast. More notably, it introduces large, abnormal orange-yellow regions in the AoLP maps around the chick's neck, indicating considerable AoLP deviations, pronounced boundary artifacts, and abrupt changes in polarization orientation.
  
SR-JCPD, an early dictionary-learning-based method, exhibits significant deficiencies in quantitative metrics. Specifically, the algorithm yields nearly the lowest PSNR and SSIM scores across both datasets, coupled with relatively high AoLP errors. In terms of visual quality, it suffers from evident polarization underestimation, dense granular artifacts in the DoLP maps, and severe texture contamination on the patterned surfaces and the chick's skin, resulting in coarse and discontinuous polarization details. Its corresponding AoLP maps exhibit rigid, unnatural variations and abrupt color transitions near structural boundaries. These phenomena indicate that this method fails to fully exploit the inherent correlation between chromatic and polarimetric information, which inevitably results in reduced overall image fidelity, distorted estimation of polarization parameters, and severe loss of fine texture details.

NLCSR achieves competitive reconstruction performance for the four polarization-intensity images, but its performance decreases substantially in terms of the PSNR of the $S_0$ image. While it effectively suppresses granular artifacts in the visual results, its DoLP maps remain undervalued, and its structural details are over-smoothed. This excessive smoothing translates to the AoLP maps, causing the ffne folds around the chick’s neck and subtle root boundaries to gradually disappear. Consequently, local polarization-orientation details and structural continuity are weakened. All these obvious reconstruction defects indicate that the algorithm inevitably yields biased DoLP values and loses subtle polarization texture details due to the failure to exploit cross-channel polarization correlation and the absence of corresponding joint constraints.

TCPDNet and PUGDiff demonstrate the effectiveness of deep-learning-based models and achieve impressive overall quantitative metrics. However, they exhibit lower SSIM values due to a bias toward pixel-level intensity accuracy over structural fidelity, which further leads to shortcomings in AoLP reconstruction — both undesirable for applications requiring high physical fidelity. Qualitatively, their AoLP results indicate that neither method adequately preserves local polarization continuity and physically meaningful boundary information. Specifically, TCPDNet tends to produce pronounced directional fluctuations and local artifacts in finely textured regions, which appear as discontinuous color transitions and localized texture contamination in the AoLP maps. Noticeable high-frequency oscillations also occur near structural boundaries, resulting in unstable polarization-orientation fields. By contrast, PUGDiff produces overly smooth results. Its DoLP and AoLP maps exhibit weakened local polarization contrast, causing fine polarization textures and boundary details to become blurred. This once again demonstrates that such methods suffer from insufficient generalization capabilities when confronted with untrained or real-world scenes. 
   
By employing a quaternion algebraic framework, the proposed method jointly exploits the intrinsic correlations and coupling relationships among the four polarization channels, thereby improving both reconstruction accuracy and physical consistency, as evidenced by its highest SSIM scores on both datasets and superior structural preservation performance. On the Guo dataset, the proposed method demonstrates overall superiority, achieving the best performance across almost all metrics. On the Wen dataset, its PSNR performance is comparable to that of deep-learning-based models, obtaining either the best or second-best results across all intensity images and $S_0$, while achieving optimal performance in terms of both SSIM and AoLP error. This further demonstrates its ability to preserve physically meaningful polarization information. The results produced by the proposed method are visually closest to the ground truth in both the DoLP and AoLP maps. Benefiting from quaternion-based modeling of the correlations among the polarization channels, the proposed method preserves sharp structural boundaries and coherent polarization distributions while effectively suppressing reconstruction artifacts. Moreover, fine polarization textures and high-frequency structural details are well retained in challenging regions, demonstrating superior polarization consistency and structural fidelity under sparse CPFA sampling.

\FloatBarrier

\section{Conclusion}
\label{sec:6}
Motivated by the substantially stronger correlations among polarization channels than among RGB channels, this work develops a quaternion tensor model to jointly encode color polarization images acquired at multiple polarization orientations. A low-rank prior is imposed on the resulting quaternion tensor to capture the global structural redundancy across the color and polarization dimensions. Moreover, the coupled gradients of the polarization-intensity images are mapped into the Stokes domain through an orthogonal transformation, thereby separating the total-intensity component from the polarization-difference components, while naturally yielding a total-intensity consistency residual. Adaptive component-wise weights encoded in quaternion form are then constructed from local gradient statistics and a global reference scale derived from the total-intensity gradient. A relatively large weight is assigned to the residual component to suppress total-intensity inconsistency, while the remaining physical components are regularized according to their respective structural characteristics.


Experimental results on two publicly available datasets demonstrate that the proposed method achieves competitive demosaicking performance in terms of both quantitative metrics and visual quality, while maintaining clear physical interpretability. In particular, as the first attempt to introduce quaternion tensor modeling into CPDM, the proposed representation provides a unified framework for jointly modeling spectral and polarization information and offers a new perspective for subsequent research on color polarization imaging.

\section*{Acknowledgments}
This work was supported in part by the National Natural Science Foundation of China under Grant Nos.~12501714 and 12561070, the Yunnan Fundamental Research Projects under Grant Nos.~202401AU070203 and 202501AT070192, the University of Macau under Grant Nos.~MYRG-GRG2024-00290-FST-UMDF and MYRG-CRG2024-00046-FHS, and in part by the China Postdoctoral Science Foundation under Grant No.~2025M783136.

\FloatBarrier
\bibliographystyle{unsrt}
\bibliography{myref}
\end{document}